\documentclass[twoside,11pt]{article}

\usepackage{blindtext}

\usepackage[preprint]{jmlr2e}

\usepackage[inline]{enumitem}
\usepackage{url}
\usepackage{comment}
\usepackage{amssymb}
\usepackage{dsfont}
\usepackage{subfig}
\usepackage{algorithm}
\usepackage{algorithmic}
\usepackage{tikz}
\usepackage{hyperref}
\usetikzlibrary{
	shapes.geometric, 
	arrows, 
	arrows.meta, 
	calc, positioning, 
	decorations.pathreplacing, 
	calligraphy
}

\makeatletter
\newtheorem{rep@theorem}{\normalfont \rep@title}
\renewcommand{\therep@theorem}{}
\newcommand{\newreptheorem}[2]{%
	\newenvironment{rep#1}[1]{%
		\def\rep@title{\textbf{#2} \ref{##1}}%
		\begin{rep@theorem}}%
		{\end{rep@theorem}}}
\makeatother
\newreptheorem{theorem}{Theorem}
\newreptheorem{lemma}{Lemma}
\newreptheorem{example}{Example}
\newreptheorem{proposition}{Proposition}

\usepackage{xcolor}

\newcommand{\clients}{\ensuremath{I}}
\newcommand{\numclients}{\ensuremath{m}}
\newcommand{\numepochs}{\ensuremath{T}}
\newcommand{\maxepochs}{\ensuremath{k}}
\newcommand{\subsetofclients}{\ensuremath{S}}
\newcommand{\client}{\ensuremath{i}}
\newcommand{\epoch}{\ensuremath{t}}
\newcommand{\gradient}{\ensuremath{\Delta}}
\newcommand{\utilityfunction}{\ensuremath{u}}
\newcommand{\eval}{\ensuremath{v}}
\newcommand{\increval}{\ensuremath{\delta\eval}}

\newcommand{\shapleyvalue}{\ensuremath{\varphi}}
\newcommand{\flmodel}{\ensuremath{F}}
\newcommand{\framework}{\ensuremath{\mathsf{FLContrib}}}
\newcommand{\validdataset}{\ensuremath{D_{\mathsf{val}}}}

\newcommand{\varepoch}{\ensuremath{z}}
\newcommand{\varepochweight}{\ensuremath{p}}

\newcommand{\varclientpos}{\ensuremath{x}}

\usepackage{amsmath}

\DeclareMathOperator*{\argmin}{arg\,min}

\usepackage{lastpage}

\ShortHeadings{History-Aware and Dynamic Client Contribution in
Federated Learning}{Ghosh et al.}
\firstpageno{1}

\begin{document}

\title{History-Aware and Dynamic Client Contribution in Federated Learning\thanks{Published at ECAI 2025.}}

\author{\name Bishwamittra Ghosh \email 	bghosh@mpi-sws.org \\
       \addr Max Planck Institute for Software Systems, Germany
       \AND
       \name Debabrota Basu\\
       \addr Univ.\ Lille, Inria, CNRS, Centrale Lille, UMR 9189 - CRIStAL, France
	   \AND
	   \name Huazhu Fu\\
	   \name Yuan Wang\\
	   \name Renuga Kanagavelu\\
	   \addr{Institute of High Performance Computing (IHPC), Agency for Science, Technology and Research (A*STAR), Singapore}
	   \AND
	   \name Jin Peng Jiang\\
	   \addr{EVYD Technology, Singapore}
	   \AND
	   \name Yong Liu\\
	   \name Rick Siow Mong Goh\\
	   \name Qingsong Wei \email 	wei\_qingsong@ihpc.a-star.edu.sg \\
	   \addr{Institute of High Performance Computing (IHPC), Agency for Science, Technology and Research (A*STAR), Singapore}
}

\editor{My editor}

\maketitle

\begin{abstract}
    	Federated Learning (FL) is a collaborative machine learning (ML) approach, where multiple clients participate in training an ML model without exposing their private data. \textit{Fair and accurate assessment of client contributions} facilitates incentive allocation in FL and encourages diverse clients to participate in a unified model training. Existing methods for contribution assessment adopts a co-operative game-theoretic concept, called Shapley value, but under \textit{restricted assumptions}, e.g., all clients' participating in all epochs or at least in one epoch of FL.

	We propose a history-aware client contribution assessment framework, called {\framework}, where client-participation is dynamic, i.e., a subset of clients participates in each epoch. The theoretical underpinning of {\framework} is based on the Markovian training process of FL. Under this setting, we directly apply the linearity property of Shapley value and compute {a historical timeline of client contributions}. Considering the possibility of a limited computational budget, we propose \textit{a two-sided fairness criteria} to schedule Shapley value computation in a subset of epochs.
	Empirically, {\framework} is efficient and consistently accurate in estimating contribution across multiple utility functions. As a practical application, we apply {\framework} to detect dishonest clients in FL based on historical Shaplee values. 	
    
\end{abstract}

\section{\textbf{Introduction}}
Over the last decade, Federated Learning (FL)~\cite{kairouz2021advances,khan2021federated,zhang2021survey} has emerged as the de facto standard for collaborative Machine Learning (ML) without exposing private data. FL facilitates the involvement of multiple data owners or local clients in  training a global ML model without the need to share any raw data~\cite{cyffers2023noisy,tavara2021federated,yin2021comprehensive}, and thus, it is aligned with the standard data protection policies like GDPR~\cite{voigt2017eu}. As the success of FL depends on the participation of a diverse pool of data contributing clients, a \textit{fair and accurate assessment of client contribution} is important to facilitate incentive allocation~\cite{yu2020fairness,lim2020hierarchical}, resolve free rider issues~\cite{fraboni2021free,zhu2021advanced}, and encourage diverse clients to participate in FL training~\cite{richardson2019rewarding,lyu2020collaborative}. Thus, assessing client contribution in FL has become a question of growing interest~\cite{wang2019measure,wang2020principled,wei2020efficient}.

\tikzstyle{nn} = [trapezium, trapezium left angle=110, trapezium right angle=110,  minimum height=0.4cm, text centered, draw=black, fill=blue!25, align=center, rotate=90]
\tikzstyle{arrow} = [->,>=stealth]
\tikzstyle{server} = [rectangle, minimum height=1cm, minimum width=1cm, draw=black]
\tikzstyle{client-selection} = [rectangle, minimum height=1cm, minimum width=1cm, draw=black]
\setlength{\textfloatsep}{2pt}
\begin{figure}[t!]
	\centering
	\scalebox{0.75}
	{
		\begin{tikzpicture}[node distance=1cm]
			
			\node [text=cyan, scale=1.28] at (-3, 0.5) {Client $ c $};
			\node [text=blue, scale=1.28] at (-3, 1.5) {Client $ b $};
			\node [text=red, scale=1.28] at (-3, 2.5) {Client $ a $};
			
			\node [client-selection, fill=cyan!50] (c1) at (0 * 3 + 0.5, 0.5) {};
			\node [client-selection, fill=white] (b1) at (0 * 3 + 0.5, 1.5) {};
			\node [client-selection, fill=red!50] (a1) at (0 * 3 + 0.5, 2.5) {};

			\node [client-selection, fill=white] (c2) at (1 * 3 + 0.5, 0.5) {};
			\node [client-selection, fill=blue!50] (b2) at (1 * 3 + 0.5, 1.5) {};
			\node [client-selection, fill=red!50] (a2) at (1 * 3 + 0.5, 2.5) {};

			\node [client-selection, fill=cyan!50] (c3) at (2 * 3 + 0.5, 0.5) {};
			\node [client-selection, fill=blue!50] (b3) at (2 * 3 + 0.5, 1.5) {};
			\node [client-selection, fill=white] (a3) at (2 * 3 + 0.5, 2.5) {};

			\node [text=black, scale=1.28] at (-3, -1) {Server};
			
			\node [server] (s1) at (0 * 3 + 2, -1) {};
			\node at (0 * 3 + 2, -1) {\includegraphics[scale=0.075]{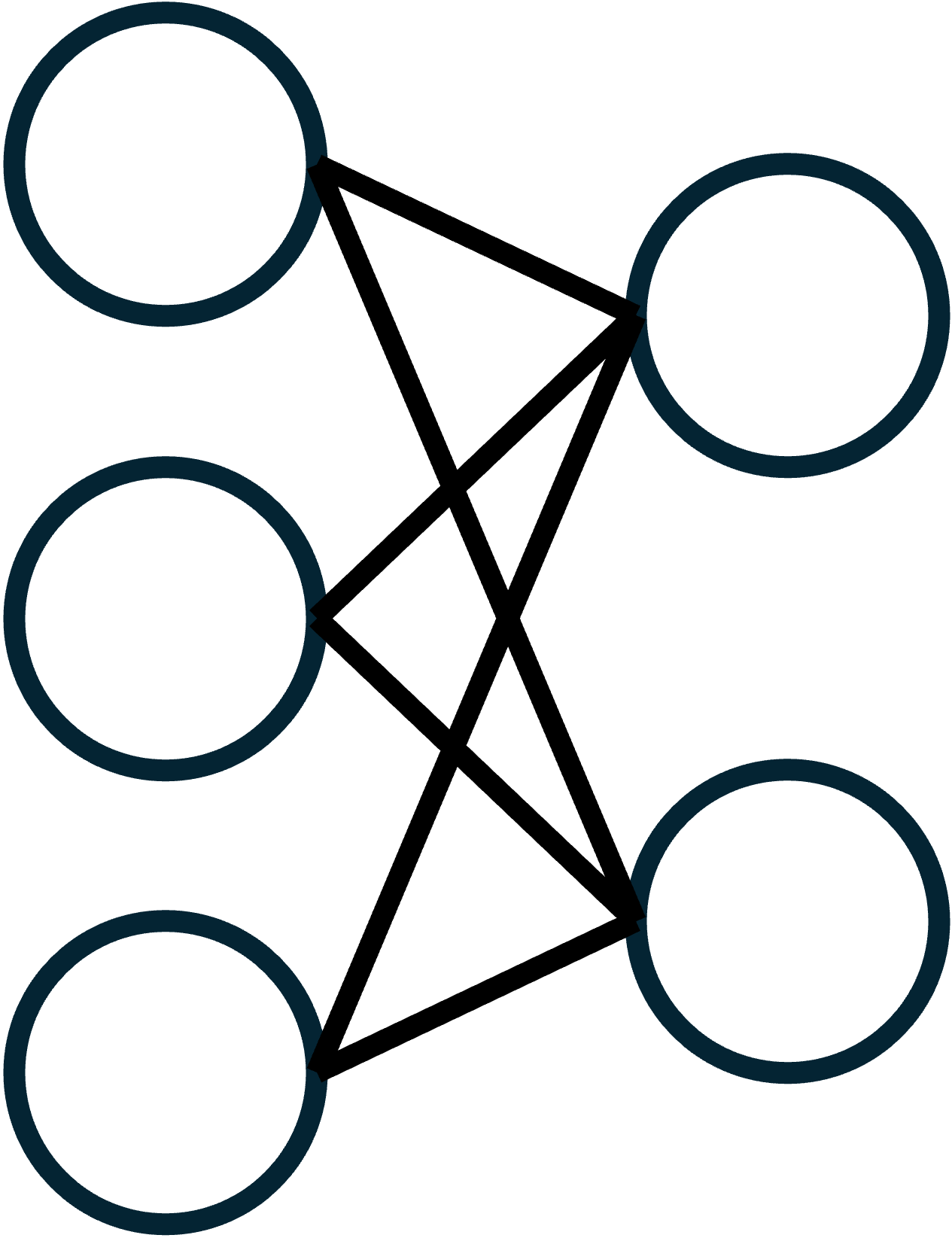}};
			\node [text=black, scale=1.28] at (0 * 3 + 2, -1.75) {$ \flmodel^{(1)} $};

			\node [server] (s2) at (1 * 3 + 2, -1) {};
			\node at (1 * 3 + 2, -1) {\includegraphics[scale=0.075]{figures/neural_network}};
			\node [text=black, scale=1.28] at (1 * 3 + 2, -1.75) {$ \flmodel^{(2)} $};
			
			\node [server] (s3) at (2 * 3 + 2, -1) {};
			\node at (2 * 3 + 2, -1) {\includegraphics[scale=0.075]{figures/neural_network}};
			\node [text=black, scale=1.28] at (2 * 3 + 2, -1.75) {$ \flmodel^{(3)} $};

			\node at (0 * 3 + 0.5, 0.5) {\includegraphics[scale=0.075]{figures/neural_network}};
			\node at (0 * 3 + 0.5, 2.5) {\includegraphics[scale=0.075]{figures/neural_network}};

			\node at (1 * 3 + 0.5, 1.5) {\includegraphics[scale=0.075]{figures/neural_network}};
			\node at (1 * 3 + 0.5, 2.5) {\includegraphics[scale=0.075]{figures/neural_network}};

			\node at (2 * 3 + 0.5, 0.5) {\includegraphics[scale=0.075]{figures/neural_network}};
			\node at (2 * 3 + 0.5, 1.5) {\includegraphics[scale=0.075]{figures/neural_network}};

			\node [text=black, text centered, align=center, scale=1.28] at (-3, -2.5) {Client\\Contribution};
			\node [text=black, text centered, align=center, scale=1.28] at (0 * 3 + 0.5, 3.25) {$ t=1 $};
			\node [text=black, text centered, align=center, scale=1.28] at (1 * 3 + 0.5, 3.25) {$ t=2 $};
			\node [text=black, text centered, align=center, scale=1.28] at (2 * 3 + 0.5, 3.25) {$ t=3 $};
			\node [text=black, scale=1.28] at (3 * 3 - 0.5, 1.5) {$ \mathbf{\cdots} $};

			\path [fill=red, draw=black]  (-1 * 3 + 1.8, -3) rectangle (-1 * 3 + 2, -2.8);
			\path [fill=blue, draw=black]  (-1 * 3 + 2, -3) rectangle (-1 * 3 + 2.2, -2.8);
			\path [fill=cyan, draw=black]  (-1 * 3 + 2.2, -3) rectangle (-1 * 3 + 2.4, -2.8);

			\node [text=black, text centered, align=center, scale=1.28] at (-1 * 3 + 2.1, -3.3) {$ t=0 $};
			\node [text=black, text centered, align=center, scale=1.28] at (0 * 3 + 2.1, -3.3) {$ t=1 $};
			\node [text=black, text centered, align=center, scale=1.28] at (1 * 3 + 2.1, -3.3) {$ t=1 $};
			\node [text=black, text centered, align=center, scale=1.28] at (2 * 3 + 2.1, -3.3) {$ t=3 $};

			\path [fill=red, draw=black]  (0 * 3 + 1.8, -3) rectangle (0 * 3 + 2, -2.5);
			\path [fill=blue, draw=black]  (0 * 3 + 2, -3) rectangle (0 * 3 + 2.2, -2.8);
			\path [fill=cyan, draw=black]  (0 * 3 + 2.2, -3) rectangle (0 * 3 + 2.4, -2.4);

			\path [fill=red, draw=black]  (1 * 3 + 1.8, -3) rectangle (1 * 3 + 2, -2.3);
			\path [fill=blue, draw=black]  (1 * 3 + 2, -3) rectangle (1 * 3 + 2.2, -2.5);
			\path [fill=cyan, draw=black]  (1 * 3 + 2.2, -3) rectangle (1 * 3 + 2.4, -2.4);

			\path [fill=red, draw=black]  (2 * 3 + 1.8, -3) rectangle (2 * 3 + 2, -2.3);
			\path [fill=blue, draw=black]  (2 * 3 + 2, -3) rectangle (2 * 3 + 2.2, -2.2);
			\path [fill=cyan, draw=black]  (2 * 3 + 2.2, -3) rectangle (2 * 3 + 2.4, -2.4);

			\draw [arrow] (s1) |- (a2.west);
			\draw [arrow] (s1) -- (b2.west);
			
			\draw [arrow] (s2) |- (b3.west);
			\draw [arrow] (s2) -- (c3.west);
			
			\draw [arrow, dashed] (a1.east) -- (s1);
			\draw [arrow, dashed] (c1.east) -- (s1);
			
			\draw [arrow, dashed] (a2.east) -- (s2);
			\draw [arrow, dashed] (b2.east) -- (s2);
			
			\draw [arrow, dashed] (b3.east) -- (s3);
			\draw [arrow, dashed] (c3.east) -- (s3);

		\end{tikzpicture}
	}	

	\caption{
		An illustration of {\framework}, where clients participate in FL dynamically (empty square indicates non-participation). The theory behind  contribution assessment is the Markovian training process of FL, where the global model update at epoch $t$ depends on local model updates by participating clients at epoch $t$, upon receiving the global model from epoch $t-1$. Thus, at any epoch, the \textit{incremental} contribution is distributed over participating clients, and non-participating clients receive null contribution. Their  \textit{total} contribution is aggregated over the history of contributions of all past epochs.}
		
	\label{fig:framework}
	\vspace{2em}
	
\end{figure}

\paragraph{Client Contribution in FL.} The existing client contribution assessment techniques in FL can be categorized in two ways: individual assessment and coalition assessment. In individual assessment, a local client is evaluated based on its similarity with the global model, the local relative accuracy, or the reputation of the client~\cite{nishio2020estimation,richardson2019rewarding}. In coalition assessment, the marginal effect of a local client is evaluated when it joins a set of other clients in the training process~\cite{ghorbani2019data,liu2022gtg,wang2019measure}. Coalition assessment is known to be more effective than individual assessment in the heterogeneous data setting, where a client only has a partial coverage of the feature space but the global model needs to cover the whole feature space~\cite{wang2020principled}. But a coalition-based assessment can depend on the order in which the clients join in the training and are being evaluated. 

\paragraph{Shapley Value as Client Contribution.} To address the client-ordering issue in coalition assessment, concepts from co-operative game theory, such as Shapley value~\cite{shapley1953value}, are proposed for assessing client contribution in FL~\cite{kang2019incentive,liu2022gtg,wang2019measure,wang2020principled}, where Shapley value determines how much a client contributes to the total utility of a coalition of clients. 
However, majority of the methods consider \textit{restricted assumptions or rely on heuristic definitions of client contribution using Shapley value}. For example,  \citet{liu2022gtg,song2019profit,wei2020efficient} considered that all clients participate in all training epochs to perform Shapley value computation, failing to simulate a real-world setting where only a subset of clients participates in training at any given epoch~\cite{fan2022improving,wang2020principled}.
To address this issue, \citet{fan2022improving} considered a non-participating-client setting, but their method of applying low rank matrix factorization incurs poor scalability and accuracy, and relies on  restricted assumptions, such as Lipschitz continuous model loss and all clients mandatorily participating together in at least one epoch.  In a similar direction,~\citet{wang2020principled} defined the contribution of non-participating clients as a heuristic, without providing any theoretical justification. Therefore, our research question considers a practical FL scenario: \textit{Can we scalably and accurately assess client contribution in federated learning, and provide a theoretical justification, when a subset of clients participates in a training epoch?}

\paragraph{Proposed Framework.} As an affirmative answer, we propose a Shapley value based framework, called {\framework}, for a history-aware and dynamic contribution assessment  of participating and non-participating clients across multiple utility functions, such as cross-entropy loss or accuracy. In particular, we resort to the Markovian training process of FL (Figure~\ref{fig:framework}), where the global model update at epoch $t$ (a new state) depends on local model updates by participating clients at epoch $t$, upon receiving the global model from epoch $t-1$ (previous state). As such, the contribution of the current epoch is distributed to participating clients in that epoch, and non-participating clients are given null contribution. Furthermore, the Markovian training process allows to directly apply the linearity property of Shapley value, where the Shapley value of a client in all epochs is the sum of incremental Shapley values of the client in individual epochs. 
In the following, we discuss our \textbf{three-fold contributions}:
\begin{itemize}
	\item \textbf{History-aware and Dynamic Client Contribution.} Due to Markovian FL training process and the application of the linearity property of Shapley value, we theoretically demonstrate that the total Shapley value of a client is the sum of incremental Shapley values in all epochs, where client-participation is dynamic. Specifically, at any epoch, the runtime complexity for computing Shapley value is dictated by participating clients only. In Lemma~\ref{lm:shapley_value_complexity}, the exact Shapley value computation is exponential with participating clients and linear with non-participating clients. 
	
	{\framework} offers a modular design, allowing us to apply off-the-shelf efficient approximation algorithms for computing Shapley value, such as via Monte Carlo sampling~\cite{castro2009polynomial} and its variants~\cite{kang2019incentive,van2018new,wang2019measure}, and also complementary contribution analysis~\cite{zhang2023efficient}. Furthermore, the separation between contribution assessment of participating and non-participating clients in {\framework} allows us to adapt existing approaches such as~\citet{liu2022gtg,wei2020efficient} focusing solely on participating clients to facilitate non-participating clients' contribution assessment.

	\item \textbf{Accuracy-efficiency Trade-off.} Computing Shapley value in all epochs leads to a more accurate assessment of client contribution, but at the cost of computational inefficiency. We simulate a trade-off between accuracy and efficiency of contribution assessment by scheduling Shapley value computation over a \textit{desirable} subset of epochs. Unlike existing methods ignoring the history of client participation~\cite{liu2022gtg,song2019profit}, we consider a two-sided fairness objectives for an optimal scheduling. Server-side fairness prioritizes epochs with higher incremental utility and higher client exposition, i.e., allowing Shapley value computation in epochs when a client participates. Client-side fairness, on the other hand, aims to minimize the pair-wise difference of exposition probability of a client with other clients. In both cases, the incremental utility of an epoch is used to prioritize Shapley value computation when global model has more impact. We express the scheduling problem as a linear program and show that {\framework} achieves an improved trade-off between accuracy and efficiency of contribution assessment.

	\item \textbf{Application: Detecting Dishonest Clients.} {\framework} yields historical client contribution over epochs as a time series of Shapley values. In a controlled experiment, we leverage such data to analyze client intention in FL training, such as identifying dishonest clients who intentionally poison their local data during training. Empirically, Shapley values via {\framework} identify the poisonous window where the client is dishonest and separate honest clients from dishonest ones.
\end{itemize}

\subsection*{Motivation and Extended Related Work}
Several approaches have applied Shapley value to assess client contribution in FL. Notably, \citet{song2019profit} utilized the gradients of local clients and proposed an one-epoch evaluation and a multi-epoch evaluation (MR) for client contribution. \citet{wei2020efficient} extended MR and proposed a truncated multi-epoch (TMR) evaluation to eliminate an entire epoch based on a pre-defined threshold. \citet{liu2022gtg} further improved computation by eliminating unnecessary epochs based on the incremental utility and proposed an improved permutation ordering of clients during assessment. \textit{In all these methods, all clients must participate in each training epoch, which we relax in our study.}

To our knowledge, two works have considered a non-participating-client setting like ours. Without any formal basis,~\citet{wang2020principled} defined the Shapley value of participating and non-participating clients at each epoch: Shapley value is initialized to $ 0 $, and then at every epoch, Shapley values of participating clients is computed using the whole history of their participations. In contrast, our analysis derives the Shapley values of all the clients exactly over a Markovian FL training and also provides the flexibility to use different initializations if one performs a warm start.
To address non-participating clients,~\citet{fan2022improving} proposed to guess the missing utility of a client-combination when a subset of clients participates in training. They relied on low rank matrix factorization to approximate the incomplete utility matrix. Their method has several limitations: low rank matrix factorization requires a suitable characteristic of the utility function, such as Lipschitz continuous model loss, and a restricted assumption on client participation, where all clients must participate together in at least one epoch. Furthermore, matrix factorization is a one-shot approach, and thus, assessing contributions in a new epoch requires re-computation of the whole matrix, which is infeasible in practice. As such,~\citet{fan2022improving} incurred high computational time and estimation error. \textit{In our study, we aim to achieve both improved accuracy and efficiency in assessing client contributions with dynamic client participation.}

\section{\textbf{Preliminaries: FL and Shapley Value}}
\textbf{Federated Learning (FL).} FL is a collaborative ML framework that allows the training of multiple clients with local private data~\cite{kairouz2021advances}. We consider a centralized single server-based FL setting, where the server is trustworthy and performs both model aggregation and contribution assessment (Figure~\ref{fig:framework}).

Let $ \clients $ denote the set of all clients with cardinality $ \numclients \triangleq |\clients| $.  $ \flmodel $ denotes the global model, and $ \numepochs $ denotes the total number of training epochs. In each epoch $ t $, $ 1 \le t \le \numepochs $, a subset of clients $ \clients^{(\epoch)} \subseteq \clients $ is selected for training. Each client $ \client \in \clients^{(\epoch)} $ receives the global model $ \flmodel^{(\epoch-1)} $ from the last epoch, trains a local model $ \flmodel^{(\epoch)}_{\client} $ on the local data $ D_{\client} $, and sends the local gradient $ \gradient^{(\epoch)}_{\client} \triangleq \flmodel^{(\epoch)}_{\client} - \flmodel^{(\epoch-1)}  $ to the server for aggregation. $\gradient^{(\epoch)}_{\client}$ may include multi-step gradients performed by the client. {\framework} is agnostic to the gradient updating algorithm.

The server preforms a federated aggregation on received local gradients $ \{\gradient^{(\epoch)}_{\client}\}_{\client \in \clients^{(\epoch)}} $ to derive a global model in each epoch. For example, the FedAvg algorithm performs the aggregation by weighing the local gradients with the relative local data size~\cite{fedavg}. In fact, FedAvg aggregates local models instead of local gradients, which is mathematically equivalent.
\begin{equation}\label{eq:fl_naive}
	\flmodel^{(\epoch)} = 	\flmodel^{(\epoch-1)} + \sum_{\client \in \clients^{(\epoch)}} \frac{|D_{\client}|}{\sum_{\client' \in \clients^{(\epoch)}}|D_{\client'}|}\Delta^{(\epoch)}_{\client}\
\end{equation}
We consider a \textit{utility function} $ \eval(\flmodel, \validdataset) \in \mathbb{R} $ to evaluate $ \flmodel $ on a hold-out validation dataset $ \validdataset $. 
The utility function can be multi-dimensional, such as model loss, accuracy, performative-fairness~\cite{zezulka2023performativity}, demographic-fairness~\cite{papadaki2022minimax}, or their combination. We use   $ \eval(\flmodel) $ to denote the utility when it is clear from the context.

\paragraph{Shapley Value.} In co-operative game theory, Shapley value~\cite{shapley1953value} computes a unique distribution of the total utility in a coalition of $ \numclients $ players. Thus, Shapley value provides the premise to assess client contributions in FL, where the total utility is distributed based on the marginal contributions of clients in all possible coalitions.

To compute Shapley value, the utility function needs to be evaluated on a FL sub-model, which is aggregated over a subset of clients' models -- our goal is to find the marginal utility when a client's model is added to the sub-model. We denote the sub-model as $ \flmodel_{\subsetofclients} $ and the utility as $ \eval(\flmodel_S, \validdataset) $, where $ \subsetofclients \subseteq \clients $. Avoiding notational clutter, we denote 
$ \utilityfunction(S) \equiv \eval(\flmodel_S, \validdataset) $ as the utility of the respective subset of clients, where $ \utilityfunction: 2^{\clients} \rightarrow \mathbb{R} $.

Therefore, Shapley value $\shapleyvalue_{\client}(\utilityfunction) $ of a client $ \client $ is defined as
\begin{equation}\label{eq:shapley_value}
	\shapleyvalue_{\client}(\utilityfunction) \triangleq \frac{1}{\numclients} \sum_{\subsetofclients \subseteq \clients \setminus \{\client\}}  \frac{1}{\binom{\numclients - 1}{|\subsetofclients|}} (\utilityfunction(\subsetofclients \cup \{\client\}) - \utilityfunction(\subsetofclients)),
\end{equation}
where $ \utilityfunction(\subsetofclients \cup \{\client\}) - \utilityfunction(\subsetofclients) $ is the marginal utility of client $ \client $ for including $\client$ to the subset of clients $ \subsetofclients \setminus \{\client\} $. Informally, Shapley value computes the average marginal utility of the client $ \client $ in all possible coalitions.  Shapley value satisfies a set of desirable properties~\cite{shapley1953value}. 
\begin{itemize}[nosep]
	\item \textbf{Decomposability (or efficiency):} The sum of Shapley values of all clients is the total utility of the global model, $ \sum_{\client \in \clients} \shapleyvalue_{\client}(\utilityfunction) = \utilityfunction(\clients) = \eval(\flmodel, \validdataset) $.
	
	\item \textbf{Symmetry:} Two clients $ \client, \client' \in \clients $ contribute equally $  \shapleyvalue_{\client}(\utilityfunction) = \shapleyvalue_{\client'}(\utilityfunction) $, if their marginal utilities are equal $ \utilityfunction(\subsetofclients \cup \{\client\}) = \utilityfunction(\subsetofclients \cup \{\client'\}) $  for every subset of clients $ \subsetofclients \subseteq \clients \setminus \{\client, \client'\} $ not containing $ \client, \client' $.

	\item \textbf{Null Client:} A null client does not contribute. The Shapley value of a null client is zero, $ \shapleyvalue_{\client}(\utilityfunction) = 0 $. This happens when the marginal utility of the client is zero, $ \utilityfunction(\subsetofclients \cup \{\client\}) - \utilityfunction(\subsetofclients) = 0 $, for every subset of clients $ \subsetofclients \in \clients \setminus \{\client\} $ not containing $ \client $.

	\item \textbf{Linearity:} Let the utility function be a linear combination of multiple utility functions. Given $ \utilityfunction_1, \utilityfunction_2 :2^{\clients} \rightarrow \mathbb{R} $, if $  \utilityfunction = \utilityfunction_1 + \utilityfunction_2 $, then the Shapley value of a client on the combined utility is a linear combination of Shapley values on individual utilities,  $ 	\shapleyvalue_{\client}(\utilityfunction) = \shapleyvalue_{\client}(\utilityfunction_1 + \utilityfunction_2) = \shapleyvalue_{\client}(\utilityfunction_1) + \shapleyvalue_{\client}(\utilityfunction_2) $.
\end{itemize}

In the paper, we apply the \textit{decomposability and null-client properties to assess contributions of participating and non-participating clients}, respectively.  In addition, we apply the \textit{linearity property to account for the Markovian training process in FL}, enabling us to perform epoch-wise contribution assessment.

\section{\textbf{Methodology}}
We present $ \framework $, a game-theoretic framework to assess history-aware client contributions in a single-server federated learning with dynamic client participation. We first formalize the problem statement and present a baseline algorithm to exactly assess client contribution with a non-participating-client setting (Section~\ref{sec:exact_client_contribution}). Then, we discuss an efficient scheduling procedure for a faster client contribution assessment (Section~\ref{sec:faster_client_contribution}). 

\textbf{Problem Statement.} Our objective is to assess contributions of clients by computing their Shapley values. Given (i) the initial global model $ \flmodel^{(0)} $, (ii) the gradients of participating clients in $ \numepochs $ training epochs $ \{\{\gradient^{(\epoch)}_{\client}\}_{\client \in \clients^{(\epoch)}}\}_{\epoch=1}^{\numepochs} $, and (iii) an utility function $ \eval $, we compute the Shapley value $ \shapleyvalue_{\client}(\eval) $ of all clients $ \client \in \clients $. Intuitively, between two clients $ \client, \client' \in \clients $, if $ \client $ has higher Shapley value than $ \client' $, $ \shapleyvalue_{\client}(\eval) > \shapleyvalue_{\client'}(\eval) $, then $ \client $ contributes more to the utility than $ \client' $.

\subsection{\textbf{Exact Client Contribution over Epochs}}\label{sec:exact_client_contribution}
To access contribution exactly, we divide key concepts into three: per-epoch Shapley value computation due to Markovian FL training, an extension of FL model update to incorporate non-participating clients, and an adaptive sub-model reconstruction to compute marginal utility. We conclude this subsection by discussing runtime complexity and error bounds.

\paragraph{Shapley Value on Incremental Utility.} FL undergoes a Markovian training process: In each epoch, the global model depends on participating clients' local models, independent of how the global model is updated in earlier epochs. As such, we directly apply the linearity of Shapley value by decomposing the total utility of the global model in all epochs as the sum of incremental utilities between consecutive epochs -- analogously, the Shapley value on the total utility is the sum of incremental Shapley values in every epoch. Most existing methods proposed to compute incremental Shapley value like ours~\cite{liu2022gtg,song2019profit,wang2020principled}, but none made the precise connection between Markovian FL training and the linearity of Shapley value.
Next, we define the incremental utility between epoch $ \epoch $ and $ \epoch - 1  $ as the difference of the model utility in consecutive epochs.

\begin{equation}
	\increval(\flmodel^{(\epoch)}, \flmodel^{(\epoch-1)}) \triangleq \eval(\flmodel^{(\epoch)}) - \eval(\flmodel^{(\epoch-1)}).
\end{equation}
\begin{lemma}[Utility Decomposition]
	\label{lm:utility_decomposition}
	The total utility in a multi-epoch FL training is the sum of incremental utilities in all the training epochs and the utility of the initial FL model.
	\begin{align}
		\eval(\flmodel^{(\epoch)}) = \sum_{\epoch = 1}^{\numepochs} \increval(\flmodel^{(\epoch)}, \flmodel^{(\epoch-1)}) + \eval(\flmodel^{(0)}),
	\end{align}
    where $\eval(\flmodel^{(0)})$ denotes the utility of the initial model.
\end{lemma}
Applying the linearity of Shapley value, the total Shapley value of a client after all epochs is the sum of Shapley values computed on the incremental utility in each epoch.
\begin{lemma}[Contribution Decomposition]
	\label{lm:shapley_value_decomposition}
	For a client $ \client $, let $ \shapleyvalue_{\client}(\eval) $ be the Shapley value after $ \numepochs $ epochs, and let $ \shapleyvalue^{(\epoch)}_{\client}(\increval) $ be the Shapley value on the incremental utility at epoch $ \epoch $. Since $ \eval $ is a linear sum of $ \increval $, we apply the linearity of Shapley value: the Shapley value w.r.t.\ $ \eval $ is the sum of Shapley values w.r.t.\ $ \increval $ between epoch $1$ to $\numepochs$ and the Shapley value of the initial model. 
	\begin{align}
		\shapleyvalue_{\client}(\eval)  = \sum_{\epoch=1}^{\numepochs} \shapleyvalue^{(\epoch)}_{\client}(\increval) + \shapleyvalue^{(0)}_{\client}(\eval)
	\end{align}
\end{lemma}

\paragraph{{Participation}-aware Model Update.} To assess contributions of both participating and non-participating clients, we express the FL model update at an epoch using all clients.

\begin{equation*}
	\begin{split}
		\flmodel^{(\epoch)} &= 	\flmodel^{(\epoch-1)} + \sum_{\client \in \clients^{(\epoch)}} w(\client, \clients^{(\epoch)})\Delta^{(\epoch)}_{\client}\nonumber\\
		&= \flmodel^{(\epoch-1)} + \sum_{\client \in \clients} \mathds{1}(\client \in \clients^{(\epoch)}) w(\client, \clients^{(\epoch)})\Delta^{(\epoch)}_{\client},\nonumber
	\end{split}
\end{equation*}
The indicator function $ \mathds{1}(\cdot) \in \{0, 1\} $ returns $ 1 $ for a true argument, such as when client $ \client $ participates at epoch $ \epoch $, and $0$ otherwise. To simplify notation, we define a \textit{participation-aware weight function} $ \lambda $ by multiplying indicator $ \mathds{1} $  and local weight $ w $, and use $\lambda$ to extend Eq.~\eqref{eq:fl_naive} to all clients. Thus, $ \lambda $ is equal to $ w $ when client $ \client $ participates, and $0$ otherwise.
\begin{align}
	\lambda(\client, \subsetofclients) &= \mathds{1}(\client \in \subsetofclients )w(\client, \subsetofclients) =
	\begin{cases}
		w(\client, \subsetofclients) \quad \text{if } \client \in \subsetofclients \\
		0 \quad \text{otherwise}
	\end{cases}\\
	\flmodel^{(\epoch)} &= \flmodel^{(\epoch-1)} + \sum_{\client \in \clients} \lambda(\client, \clients^{(\epoch)})\Delta^{(\epoch)}_{\client}\label{eq:fl_all_clients}
\end{align}
\begin{example}
	\label{ex:lambda_function}
	\normalfont
	In Figure~\ref{fig:framework}, let three clients be $ \clients =  \{a, b, c\} $ with $ |D_a| = |D_b| = |D_c| $ .  In epoch $ \epoch = 1 $, $ \clients^{(1)} = \{a, c\} $. Hence, $  \lambda(a, \clients^{(1)}) = \lambda(c, \clients^{(1)}) = \frac{1}{2}  $ and $ \lambda(b, \clients^{(1)}) = 0 $. 
\end{example}

\paragraph{Adaptive Sub-model Reconstruction.} An intermediate step for computing Shapley value is to compute the marginal utility of a client w.r.t. its presence and absence in a subset of other clients. Instead of retraining on all possible subsets, we \textit{reconstruct} a sub-model by storing and utilizing local gradients of participating clients at the FL server, similarly as ~\citet{liu2022gtg,song2019profit}. Importantly, we consider an \textit{adaptive weighting} of local gradients  to ensure that each sub-model reconstruction mimics the full global model training. In particular, the weight is defined as $ w(\client, \subsetofclients) \triangleq |D_\client| / (\sum_{\client' \in \subsetofclients}|D_{\client'}|) $, which is adaptive w.r.t. $\subsetofclients$. The adaptive weighting is overlooked in earlier methods by~\citet{lin2023fair}, where $ w(\client, \subsetofclients) = |D_\client| / (\sum_{\client' \in \clients}|D_{\client'}|) $, ignoring the effect of $ \subsetofclients $, and naively normalizing for all clients $ \clients $.
We incorporate this weight into the participation-aware weight function $ \lambda $ for reconstructing the sub-model $ \flmodel^{(\epoch)}_{\subsetofclients} $.

\begin{equation}
	\flmodel^{(\epoch)}_{\subsetofclients} 
	= \flmodel^{(\epoch-1)} + \sum_{\client \in \subsetofclients} \lambda(\client, \subsetofclients \cap \clients^{(\epoch)})\Delta^{(\epoch)}_{\client}
	\label{eq:fl_sub_model}
\end{equation}

Here, $\lambda(\client, \subsetofclients \cap \clients^{(\epoch)})$ ensures that only participating clients in $ \subsetofclients \cap \clients^{(\epoch)} $ receive non-zero weights, i.e., their Shapley value is computed in the traditional way, since they explicitly update the global model at the current epoch. In contrast, non-participating clients do not influence sub-model reconstruction (Lemma~\ref{lm:non_selected_submodel}) and thereby receive a deterministic Shapley value (Lemma~\ref{lm:shapley_value_non_selected}).

\begin{lemma}[Influence of Non-participating clients]
	\label{lm:non_selected_submodel}
	Let $ \flmodel^{(\epoch)}_{\subsetofclients} $ be a sub-model consisting of a subset of clients $ \subsetofclients \subseteq \clients $ and $ \client \notin \clients^{(\epoch)} $ be a non-participating client in epoch $ \epoch $. Non-parcipating clients do not influence sub-model reconstruction, formally
	$ \flmodel^{(\epoch)}_{\subsetofclients \cup \{\client\}} = \flmodel^{(\epoch)}_{\subsetofclients} $.
\end{lemma}

\begin{example}[Continuing Example~\ref{ex:lambda_function}]
	\normalfont
	\label{ex:adaptive_sub_model}
	Since $ \clients^{(1)} = \{a, c\} $, then client $ b $ does not participate in epoch $ \epoch = 1 $. Hence, $ \flmodel^{(1)}_{\{a, b, c\}} = \flmodel^{(1)}_{\{a, c\}} 
	= \flmodel^{(0)} + \frac{1}{2} \Delta^{(1)}_{a} + \frac{1}{2} \Delta^{(1)}_{c} $, $ \flmodel^{(1)}_{\{a, b\}} = \flmodel^{(1)}_{\{a\}} 
	= \flmodel^{(0)} +  \Delta^{(1)}_{a}, \flmodel^{(1)}_{\{b, c\}}  = \flmodel^{(1)}_{\{c\}} 
	= \flmodel^{(0)} +  \Delta^{(1)}_{c} $ and $ \flmodel^{(1)}_{\{b\}}  = \flmodel^{(1)}_{\emptyset} 
	= \flmodel^{(0)} $.
\end{example}

\begin{lemma}[Deterministic Contribution]
	\label{lm:shapley_value_non_selected}
	At epoch $ t $, the Shapley value of a non-participating client $\client \notin \clients^{(\epoch)} $ with respect to the incremental utility $ \increval $ is zero, $ \shapleyvalue^{(t)}_{\client}(\increval) = 0 $. Therefore, the non-participating client is a null client.
\end{lemma}

\citet{wang2020principled} \textit{heuristically} defined the same deterministic Shapley value of non-participating clients, whereas our paper provides necessary theoretical justification.

\begin{remark}[Initial FL model and Shapley Value] 
	\normalfont
	We explicate the need to compute Shapley value of the initial FL model, which can be a randomly initialized model or a model after warm start. Hence, the utility of the initial FL model $ \flmodel^{(0)} $ is not necessarily zero, rather the utility of $ \flmodel^{(0)} $ on the validation dataset $ \validdataset $. Thus, a reasonable initial Shapley value is the per-client utility of the initial model, $ \shapleyvalue^{(0)}_{\client}(\eval) = \eval(\flmodel^{(0)})/|\clients| $.
\end{remark}

\begin{algorithm}[tb]
	\caption{Assessing Client Contribution}
	\label{alg:framework}
	\begin{algorithmic}[1]
		\STATE {\bfseries Input:} Initial global model $ \flmodel^{(0)} $, clients $ I $, utility $ \eval $.
		\STATE  $\shapleyvalue^{(0)}_{\client} = \frac{\eval(\flmodel^{(0)})}{|\clients|}, \forall \client \in \clients$
		\FOR{$\epoch=1$ {\bfseries to} $\numepochs$}
		\STATE $ \clients^{(\epoch)} \gets \mathtt{ClientSelection}(\clients) $
		\STATE $ \gradient^{(\epoch)}_{\client} \gets \mathtt{ClientTraining}(\flmodel^{(t-1)}, D_{\client}), \forall \client \in \clients^{(\epoch)} $
		\STATE $ \flmodel^{(t)} \gets \mathtt{FedAggregate}(\flmodel^{(t-1)}, \{\gradient^{(\epoch)}_{\client}\}_{\client \in \clients^{(\epoch)}}) $
		\STATE $ \{\shapleyvalue^{(\epoch)}_{\client} \} \gets \mathtt{FLContrib}( \increval, \{\gradient^{(\epoch)}_\client\}, \flmodel^{(\epoch-1)})$ 
		\ENDFOR
		\STATE {\bfseries return} $ \{\sum_{\epoch=0}^{\numepochs}\shapleyvalue^{(\epoch)}_{\client}\} $
	\end{algorithmic}
\end{algorithm}

\paragraph{Runtime Complexity For Exact Computation.} Due to the deterministic Shapley value of non-participating clients, the runtime complexity is linear for non-participating clients and exponential for participating clients. Non-participating clients incur a linear complexity since their contribution is recorded as a history for later use. The exponential complexity of participating clients is a known result for an exact Shapley value computation.

\begin{lemma}[Runtime Complexity]
	\label{lm:shapley_value_complexity}
	Let $ \numclients $ be the total number of clients and $ \frac{1}{\tau} \in [0,1] $ be the ratio of participating to non-participating clients. In an epoch, the runtime complexity of exactly computing Shapley value is $ \mathcal{O}(2^{\frac{\numclients}{\tau}} + (1 - \frac{1}{\tau})\numclients) $. In $ \numepochs $ epochs, the total running time is $ \mathcal{O}(2^{\frac{\numclients}{\tau}}\numepochs + (1 - \frac{1}{\tau})\numclients\numepochs) $.
\end{lemma}

\paragraph{Modular Design and Approximation Error.}
	\label{rm:modular_design}
	{\framework} has a modular design, where a wide variety of Shapley value algorithms as well as existing contribution methods in FL can be applied (ref.\ Algorithm~\ref{alg:framework}). The exact computation can be replaced with approximation algorithms, such as Monte Carlo sampling~\cite{ghorbani2019data} or complementary contribution analysis~\cite{zhang2023efficient}. Also, existing methods by~\citet{liu2022gtg,wei2020efficient} can be applied, where non-participating-client setting is ignored: In each epoch, we apply their Shapley value computation algorithm to participating clients and compute a deterministic Shapley value for non-participating clients. Below, we provide the error bound of {\framework} as a meta approach invoking existing Shapley value algorithms.

\begin{lemma}[Error Bound]
	\label{lm:error_bound}
	If we apply an $\epsilon$-approximation algorithm to compute the incremental Shapley values at each epoch, the total estimation error in the global Shapley value is $\mathcal{O}(\frac{\numepochs\epsilon}{\tau})$, which is of the same order as existing algorithms.
\end{lemma}

\subsection{\textbf{A Scheduler for Faster Computation}}
\label{sec:faster_client_contribution}
We discuss a scheduling procedure to assess client contribution efficiently without sacrificing accuracy. We consider a computational budget, such as the maximum number of epochs to compute Shapley value, and derive an optimal subset of epochs by solving a constrained optimization problem.

We propose two-fold objectives in the optimization problem to incorporate the history of client participation. \textbf{Server-sided fairness}: The server aims to increase utility coverage by computing Shapley value in a desirable subset of epochs where the sum of incremental utilities is higher. Also, the server prioritizes higher exposure of clients: informally, a client is said to be exposed if it participates in training in an epoch and Shapley value is computed in the same epoch. \textbf{Client-sided fairness}:
Each client aims to minimize the pair-wise difference of exposure probabilities with other clients.
Now, we discuss two mixed integer linear programs (MILP) to achieve (i) only server-sided fairness and (ii) two-sided fairness~\cite{do2021two}.

\paragraph{An ILP for One-sided Fairness.} Let $ \varepoch^{(\epoch)} \in \{0, 1\} $ be a binary variable indicating whether  Shapley value is computed in epoch $ \epoch $. The epoch weight $ \varepochweight^{(\epoch)} $ is defined as normalized absolute incremental utility, $ \varepochweight^{(\epoch)} =  \frac{|\increval(\flmodel^{(\epoch)}, \flmodel^{(\epoch-1)})|}{\sum_{\epoch=1}^{\numepochs}|\increval(\flmodel^{(\epoch)}, \flmodel^{(\epoch-1)})| + |\eval(\flmodel^{(0)})|} $, and $ \sum_{\epoch=1}^{\numepochs} \varepochweight^{(\epoch)} = 1 $. For each client, we first compute per-epoch participation rate, $ \varclientpos^{(\epoch)}_{\client} = \mathds{1}(\client \in \clients^{(\epoch)})/\sum_{\epoch=1}^{\numepochs}\mathds{1}(\client \in \clients^{(\epoch)}) $. Hence, the exposure probability of a client at epoch $ \epoch $ is defined as  $ \varclientpos^{(\epoch)}_{\client} \varepoch^{(\epoch)} $. Let $ \maxepochs \le \numepochs $ denote the maximum number of epochs to compute Shapley value. The following ILP program satisfies server-sided fairness.
\begin{align}
	\max_{\mathbf{z}}& 
	\sum_{\epoch = 1}^{\numepochs} \varepochweight^{(\epoch)} \varepoch^{(\epoch)}  +  \gamma \sum_{\client \in \clients} \sum_{\epoch=1}^{\numepochs} \varclientpos^{(\epoch)}_{\client} \varepoch^{(\epoch)} \label{eq:server_sided_fairness} \\ 
	=& \sum_{\epoch = 1}^{\numepochs} \Big( \varepochweight^{(\epoch)} + \gamma  \sum_{\client \in \clients} \varclientpos^{(\epoch)}_{\client}   \Big) \varepoch^{(\epoch)} \notag \\
	\text{such that} &\sum_{\epoch = 1}^{\numepochs} \varepoch^{(\epoch)} \le  \maxepochs \notag
\end{align}

In the maximization problem, we jointly maximize the utility coverage of epochs (the first term) and the exposure probability of clients (the second term), both of which are linear with  $ \varepoch^{(\epoch)} $. We put a constraint that the sum of $ \varepoch^{(\epoch)} $ is at most $ \maxepochs $ to meet the budget constraint. In addition, we consider a hyper-parameter $ \gamma \in \mathbb{R}^{\ge 0} $ to prioritize between utility coverage and client exposition. An off-the-shelf ILP solver can output an optimal solution of $ {\mathbf{z}^{(\epoch)}}^* $ such that we compute Shapley value in an epoch $ \epoch $ if and only if $ {\varepoch^{(\epoch)}}^* = 1 $.

\paragraph{An MILP  for Two-sided Fairness.} To achieve server-client two-sided fairness, we consider a joint optimization problem: we maximize the utility coverage of epochs where Shapley value is computed and minimize the pair-wise absolute exposure probability of clients in Eq.~\eqref{eq:two_sided_fairness}.

\begin{align}
	\max_{\mathbf{z}}&\sum_{\epoch = 1}^{\numepochs} \varepochweight^{(\epoch)} \varepoch^{(\epoch)}  -  \gamma \sum_{\client, \client' \in \clients} |\sum_{\epoch=1}^{\numepochs} (\varclientpos^{(\epoch)}_{\client} -  \varclientpos^{(\epoch)}_{\client'}) \varepoch^{(\epoch)}| 	\label{eq:two_sided_fairness} \\
	\text{such that } &\sum_{\epoch = 1}^{\numepochs} \varepoch^{(\epoch)} \le  \maxepochs  \notag
\end{align}

Since the objective function has absolute terms, a direct reduction to an MILP program requires defining $ \mathcal{O}(\numclients^2) $ auxiliary real-valued variables and linear constraints.

\textbf{An alternate approach} is to maximize the lower bound of the objective function in Eq.~\eqref{eq:two_sided_fairness}, which asks for an ILP solution with respect to the variable $ \varepoch^{(\epoch)} $. 
\begin{align}
	&\sum_{\epoch = 1}^{\numepochs} \varepochweight^{(\epoch)} \varepoch^{(\epoch)}  -  \gamma \sum_{\client, \client' \in \clients} |\sum_{\epoch=1}^{\numepochs} (\varclientpos^{(\epoch)}_{\client} -  \varclientpos^{(\epoch)}_{\client'}) \varepoch^{(\epoch)}|\notag\\
	\geq&\sum_{\epoch = 1}^{\numepochs} (\varepochweight^{(\epoch)}   -  \gamma  \sum_{\client, \client' \in \clients} | (\varclientpos^{(\epoch)}_{\client} -  \varclientpos^{(\epoch)}_{\client'})|) \varepoch^{(\epoch)}\label{eq:two_sided_fairness_lower_bound}
\end{align}

Since in the alternate approach we maximize the lower bound of the objective function, the approach is more restrictive than the former approach. Here, we refer to it as \textbf{Two-sided Fair LB} scheduling.

\begin{remark} 
	\normalfont
	We explain cases when one-sided versus two-sided fairness is preferred. Let us consider two kinds of non-uniformity: (1) some epochs have more participating clients than other epochs, i.e., non-uniform client-participation across epochs, (2) some clients participate more than others, i.e., non-uniform participation across clients. For (1), one-sided fairness is preferred where epochs with higher participation are selected to increase client exposure, i.e., when more clients participate, Shapley value is computed for a granular contribution assessment. For (2), two-sided fairness is preferred where the pair-wise difference of client exposure is minimized, i.e., one client is not favored more than another client to compute Shapley value. Finally, for the trivial case when all clients participate uniformly across epochs, both one-sided and two-sided fairness may act  equivalently, where one can simply rank epochs by corresponding incremental utilities, and choose the top $ \maxepochs $ epochs to compute Shapley value without needing to solve an ILP program.

\end{remark}

\begin{lemma}[Continuing Lemma~\ref{lm:shapley_value_complexity}]
	\label{lm:complexity_scheduling}
	For $ \maxepochs \le \numepochs $ denoting the maximum number of epochs for Shapley value computation, the runtime complexity is $ \mathcal{O}(2^{\frac{\numclients}{\tau}}\maxepochs + (1 - \frac{1}{\tau})\numclients\maxepochs) $, which is $\frac{\maxepochs}{\numepochs}$ fraction of the total runtime complexity without scheduling.
\end{lemma}

\section{\textbf{Empirical Performance Evaluation}}

We conduct an empirical evaluation of {\framework}. In the following, we discuss  objectives of the experiments, experimental setup, and experimental results.  \textbf{Our objectives of experiments} are two-fold\footnote{Theoretical proofs, details on experimental setup such as FL training, and additional experimental results including model aggreegation using Shapley value, the impact of different approximation algorithms for Shapley value etc.\ are in the Appendix.}.

\begin{itemize}
	\item \textit{Comparative Performance:} How does {\framework} compare with existing methods in computational time and estimation error of client contribution assessment?
	\item \textit{Ablation Study:} How do different scheduling procedures and associated parameters in {\framework} impact the computational time and estimation error of client contribution assessment?
\end{itemize}

\paragraph{Experimental Setup.} We implement a prototype of {\framework} in Python $ 3.8 $. For Shapley value computation, we consider an approximation algorithm based on complementary contribution~\cite{zhang2023efficient}. For a faster  scheduling (Section~\ref{sec:faster_client_contribution}), we consider model loss as the epoch-weight $ \varepochweight^{(\epoch)} $, while assessing client contribution to mutliple utility functions such as loss and accuracy.  In addition, we normalize each term in the objective functions in Eq.~\eqref{eq:server_sided_fairness},~\eqref{eq:two_sided_fairness},~\eqref{eq:two_sided_fairness_lower_bound} and consider $ \gamma=1 $ for an equal priority. We compare {\framework} with existing methods, namely ComFedSV~\cite{fan2022improving}, GTG-Shapley~\cite{liu2022gtg}, and TMR~\cite{wei2020efficient}. Among them, we \textit{adapt} GTG-Shapley and TMR to the \textit{non-participating-client} setting. We consider three datasets: Adult~\cite{misc_adult_2}, COMPAS~\cite{angwin2016machine}, and CIFAR10~\cite{Krizhevsky09learningmultiple}, where the first two datasets are on tabular data and the last one is on image data. We experiment with two types of neural networks: a fully connected MLP model for tabular dataset and a CNN model for image classification. In comparative evaluation, we consider 132 benchmark instances for assessing client contribution by varying the number of clients in $\{4, 8, 16, 32, 64\}$, the number of training epochs in $\{12, 25, 37, 50\}$, and three random seeds. We consider a cut-off time for contribution assessment to $ 12 $K seconds. In the following, we discuss our results.

\begin{figure*}
	\centering
	\includegraphics[scale=0.4]{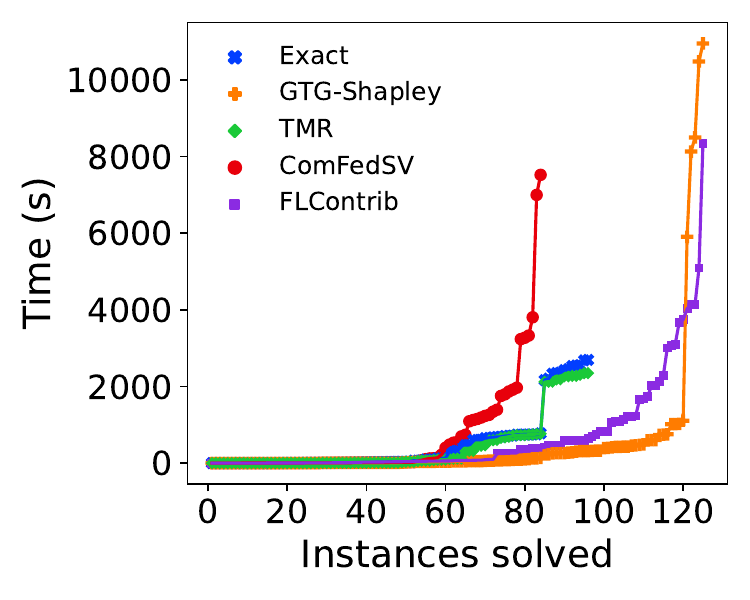}\hfil
	\includegraphics[scale=0.4]{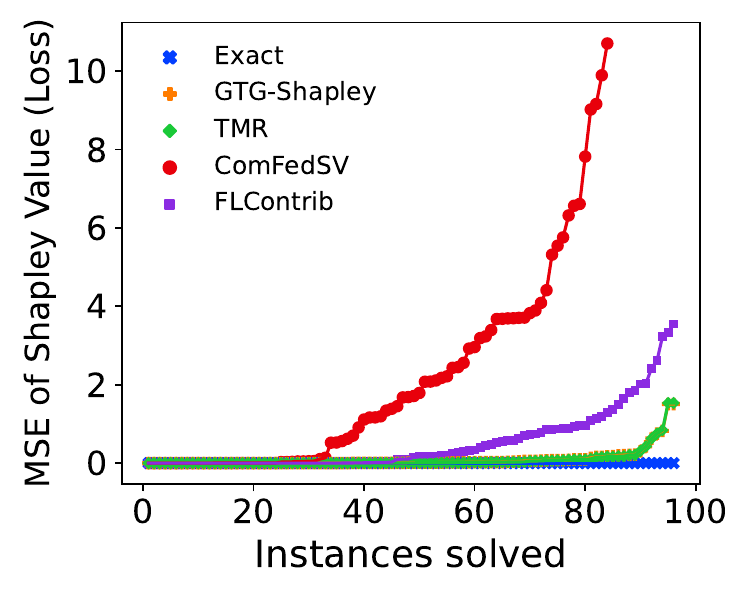}\hfil
	\includegraphics[scale=0.4]{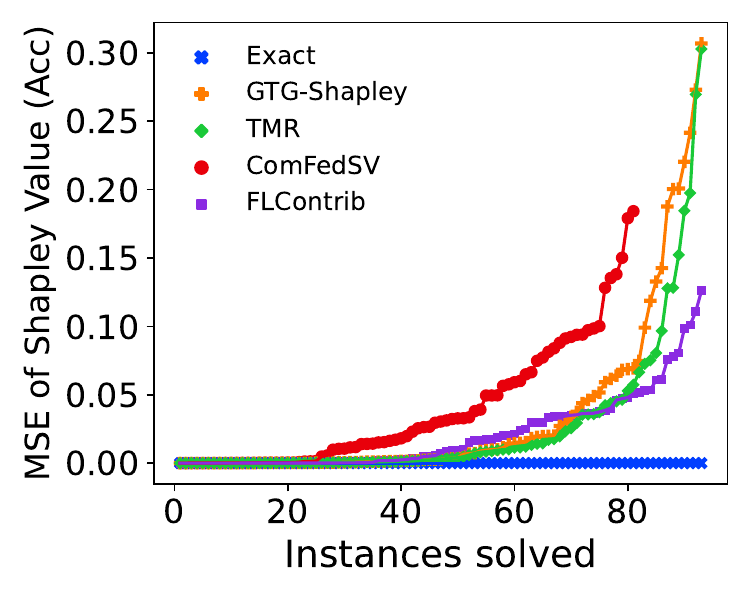}
	\caption{
		Comparison of different methods for assessing client contribution in FL based on computational time (left), and mean squared error or MSE of Shapley value on model loss (middle) and model accuracy (right).  In each cactus plot, a point $ (x, y) $ denotes that a method can assess contribution in $ x $ many instances within $ y $ seconds (left), or within $ y $ MSE (middle and right). Therefore, any method reaching the bottom-right part of the plot has the best performance. {\framework} is the most efficient method in computational time, while being consistently accurate in contribution assessment across multiple utility functions: model loss and accuracy.}\label{fig:all_methods}
		\vspace{2em}
\end{figure*}

\paragraph{Results of Computational Efficiency: {\framework} is the Most Efficient Method.} We compare the computational time of different methods in Figure~\ref{fig:all_methods} (left). We consider $ 132 $ benchmark instances, where each instance implies an FL training on three datasets with different number of clients, epochs, and repeated runs. Firstly, the exact approach (based on Section~\ref{sec:exact_client_contribution}) and TMR solve  $ 96 $ instances each and  time out in rest of the instances. Secondly, ComFedSV, the state-of-the-art for contribution assessment in the non-participating client setting, solves only $ 84 $ instances. Finally,  {\framework} with two-sided fairness and GTG-Shapley solve $ 125 $ instances each, with {\framework} taking a maximum of $ 8.3  $K seconds vs.\ $ 10.9 $K seconds by GTG-Shapley. \textit{Therefore, {\framework} is the most computationally efficient method in client contribution assessment than the existing methods.}

\paragraph{Results of Estimation Error: {\framework} is Consistently Accurate across Multiple Utility Functions.} We compare the mean squared error (MSE) of Shapley value estimated by different methods and report results in Figure~\ref{fig:all_methods} (middle and right). We consider $ 96 $ instances where the exact method computes Shapley value within the cut-off time. Firstly, ComFedSV performs poorly by incurring higher MSE than others. Secondly, GTG-Shapley and TMR are tailored for model loss as the utility function and achieve lower MSE than {\framework} (Figure~\ref{fig:all_methods} middle) -- both approaches incur similar estimation error and their curves overlap. However, when considering model accuracy as utility, they incur higher MSE than {\framework} (Figure~\ref{fig:all_methods} right). Importantly, {\framework} yields consistent estimation error across different utility functions. \textit{Therefore, {\framework} achieves well-balanced performance in estimation error than all competitive methods.}

\begin{figure}
	\centering
	
	\includegraphics[scale=0.4]{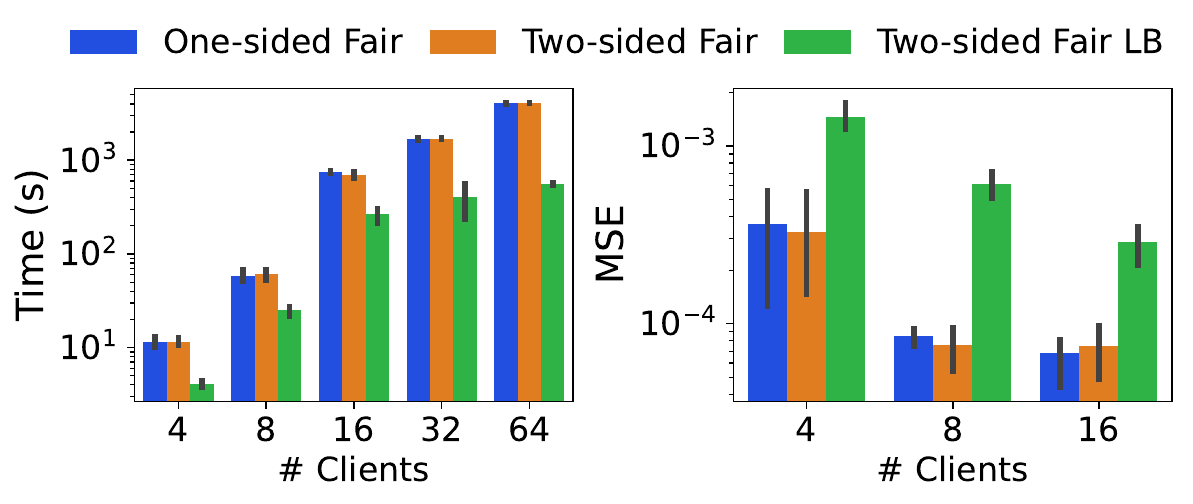}

	\includegraphics[scale=0.4]{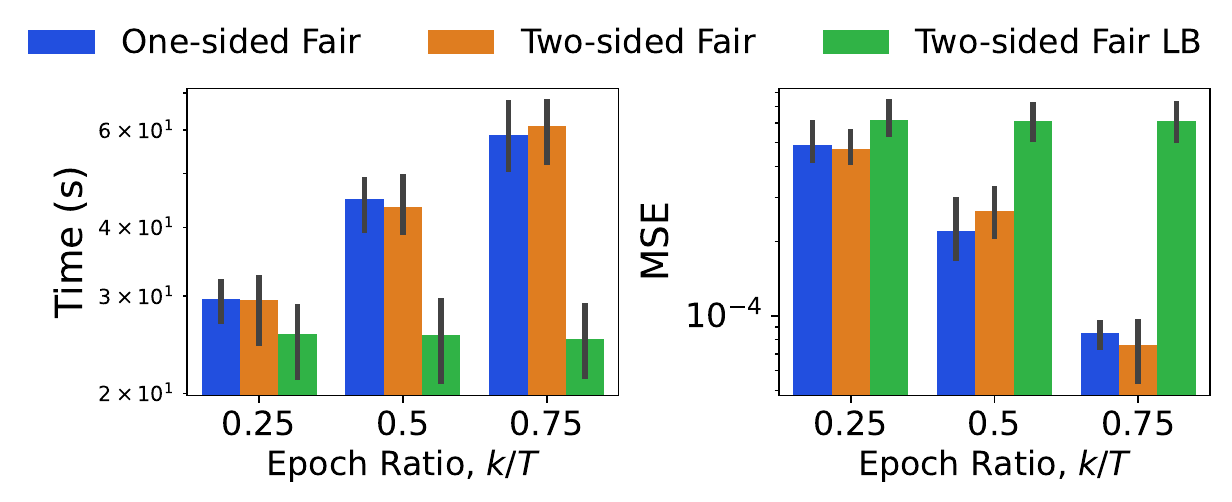}

	\caption{
		Effect of the number of clients (top) and the maximum epochs parameter in different scheduling procedures (bottom) on computation time and MSE of Shapley values.}\label{fig:ablation_study}

	\vspace{2em}
\end{figure}

\paragraph{Ablation Study.} We analyze the impact of different scheduling procedures in {\framework} while varying the number of clients in Figure~\ref{fig:ablation_study} (top) and maximum epoch for Shapley value computation in Figure~\ref{fig:ablation_study} (bottom). 

In Figure~\ref{fig:ablation_study} (top), increasing clients results in higher computational time and less estimation error of Shapley value by different scheduling procedures in {\framework}. While it is expected that with more clients, computational time will increase, the decrease in error of Shapley value is less intuitive, and observed across different scheduling procedures and parameter settings. In addition, {\framework} with one-sided and two-sided fairness demonstrate a similar performance  in both computational time and MSE. However, {\framework} with two-sided fair LB solves a more restrictive objective function and results in lower computational time but higher MSE than others.

In Figure~\ref{fig:ablation_study} (bottom), we vary the ratio $\frac{\maxepochs}{\numepochs}$ -- a higher ratio denotes computing Shapley value in more epochs. As a result, the computation time increases with an increase in the ratio, resulting in lower MSE -- the pattern is observed across different scheduling procedures. Therefore, the parameter $\maxepochs$ effectively controls the trade-off between computational time and estimation error in {\framework}.

\paragraph{Summary of Results.} {\framework} demonstrates an efficient performance in computation time while achieving a consistent estimation error of Shapley value in multiple utility functions. In particular, out of $ 132 $ instances, {\framework} solves $ 125 $ instances within the cut-off time and yields the lowest estimation error of Shapley value w.r.t.\ model accuracy. In ablation study, different parameters and scheduling procedures of {\framework} precisely control the trade-off between accuracy and efficiency of contribution assessment in FL.

\section{\textbf{Applications of {\framework}}}
We demonstrate the application of {\framework} in analyzing client intention based on historical client contributions over multiple epochs. Unlike~\cite{fan2022improving,wang2020principled,liu2022gtg,song2019profit}, we consider the history of client contributions rather than a single contribution value post training, since historical contribution is more informative to detect sudden behavioral change in individual clients. We consider a controlled experimental setup: each client possesses a uniform data distribution, except a subset of \textit{dishonest clients} who intentionally poison their data within a window of several epochs -- it is unknown beforehand when clients switch their intention. For illustration, a dishonest client chooses to flip the data label with a certain flipping probability inside the window.

\paragraph{Objectives.} We consider two objectives: (i) Can we detect the window in which  a dishonest client poisons its data? (ii) Can we separate honest clients from dishonest ones? In both cases, we rely on the cumulative Shapley value of clients over epochs and answer affirmatively to the two questions.

\begin{figure*}
	\centering
	
	\subfloat[]{\includegraphics[scale=0.4]{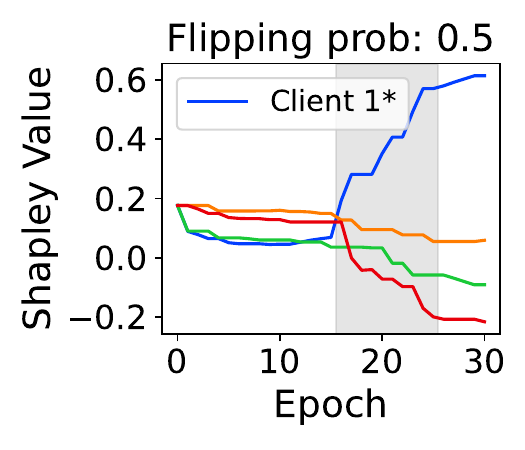}\label{fig:change_point_detection_a}}
	\subfloat[]{\includegraphics[scale=0.4]{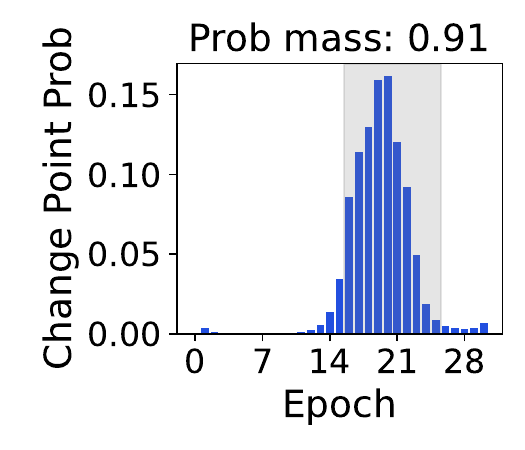}\label{fig:change_point_detection_b}}
	\subfloat[]{\includegraphics[scale=0.4]{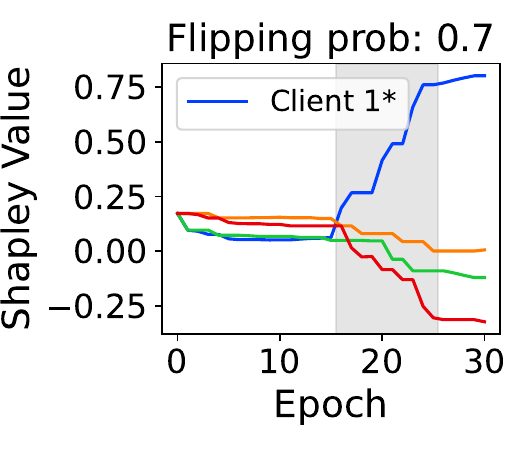}\label{fig:change_point_detection_c}}
	\subfloat[]{\includegraphics[scale=0.4]{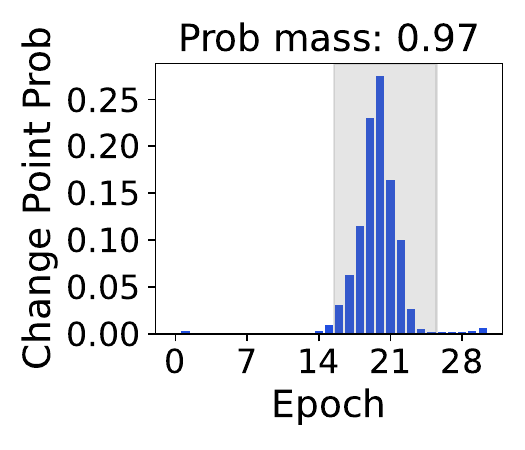}\label{fig:change_point_detection_d}}
	
	\subfloat[]{\includegraphics[scale=0.4]{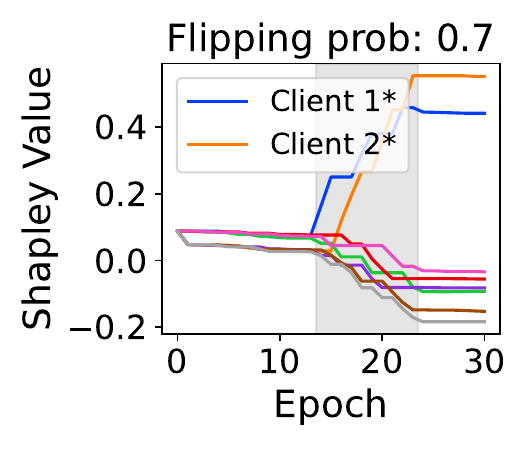}\label{fig:change_point_detection_e}}
	\subfloat[]{\includegraphics[scale=0.4]{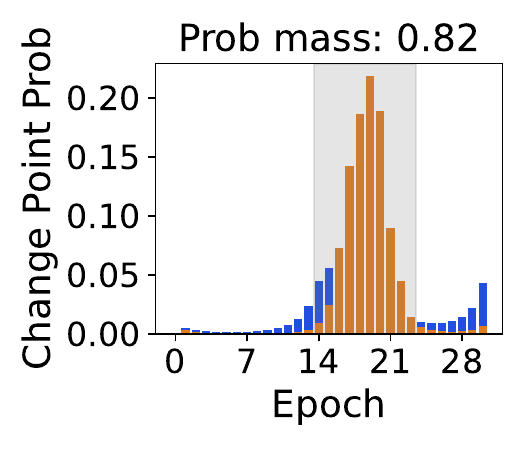}\label{fig:change_point_detection_f}}
	\caption{
		Cumulative Shapley value of individual clients computed on global model loss over multiple epochs (Figure~\ref{fig:change_point_detection_a},~\ref{fig:change_point_detection_c}, and~\ref{fig:change_point_detection_e}). We simulate a poisonous window when a subset of clients (marked with *) acts dishonestly by poisoning their local data and adversely impacting global loss -- \textit{historical Shapley value can detect this event}, where Shapley value of dishonest clients increases suddenly and substantially than honest clients. Consequently, we compute the change point detection probability of Shapley values in Figure~\ref{fig:change_point_detection_b}, ~\ref{fig:change_point_detection_d}, and~\ref{fig:change_point_detection_f}, respectively, identifying the poisonous window shown in the gray color.}\label{fig:change_point_detection}
\end{figure*}

\paragraph{Detecting Poisonous Window.} In Figure~\ref{fig:change_point_detection_a}, we consider $ 1 $ out of $ 4 $ clients as dishonest, where the cumulative Shapley value on model loss of the dishonest client diverges from honest clients within the poisonous window (gray colored region). In Figure~\ref{fig:change_point_detection_b}, we apply a Bayesian inference based change point detection algorithm~\cite{fearnhead2006exact} with the goal of identifying the poisonous window -- the probability of change point of Shapley values has significantly higher mass within the window (average probability mass  is $ 0.91 $), showing the potential of applying any threshold-based classifier for the purpose of detection. Also, increasing the flipping probability from $ 0.5 $ in Figure~\ref{fig:change_point_detection_a} to $ 0.7 $ in Figure~\ref{fig:change_point_detection_c}, i.e., intensifying the level of dishonestly, the probability mass increases from $ 0.91 $ to $ 0.97 $ -- a greater chance of detection. In another dimension, we increase total clients to $ 8 $ and dishonest clients to $ 2 $ in Figure~\ref{fig:change_point_detection_e}, where the change point probability becomes $ 0.82 $, which is still evident of poisonous intention. \textit{All these evidences support our claim that historical client contribution via Shapely value has the potential of being an effective identification of poisonous client intention in FL.}

\begin{figure}
	\centering
	\includegraphics[scale=0.4]{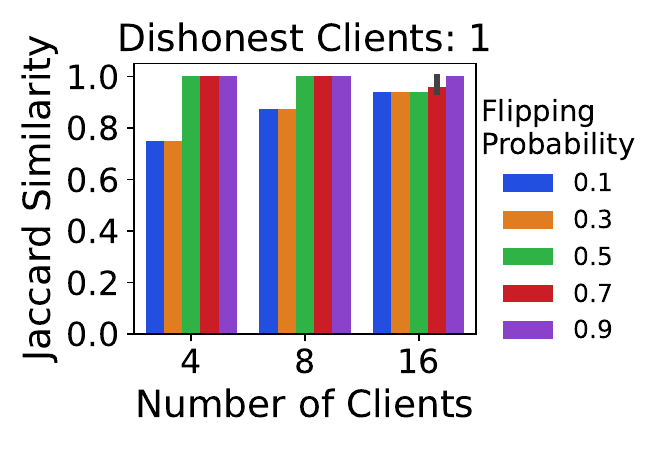}
	\includegraphics[scale=0.4]{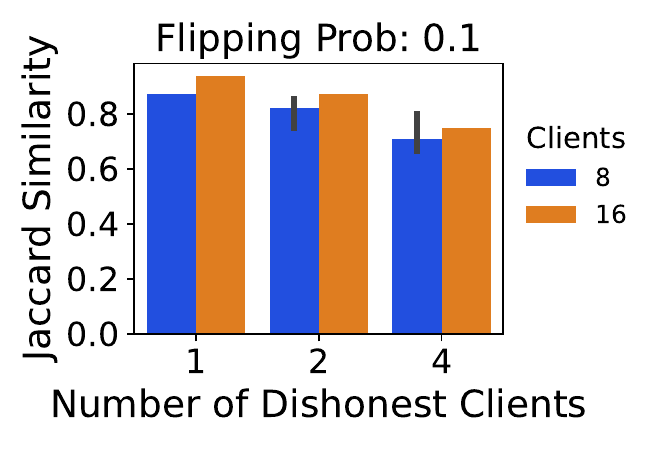}
	\caption{
		Effectiveness of separating honest clients from dishonest ones based on historical client contribution -- a higher Jaccard similarity denotes a higher separation. Separation success increases when the flipping probability or poisonous intention of dishonest clients increases, and when total clients increases while keeping dishonest clients fixed. In contrast, separation success decreases as more clients become dishonest w.r.t.\ fixed total clients.}\label{fig:jaccard_similarity}
	\vspace{2em}
\end{figure}

\paragraph{Separating Honest Clients.} Historical contribution of clients capture their intention during FL training. As such, to separate honest clients from dishonest ones, we apply K-means time-series clustering algorithm~\cite{JMLR:v21:20-091} on the cumulative Shapley value of clients. Our hypothesis is that majority of the  honest clients should be in one cluster due to their uniform data distribution, while placing dishonest clients in separate clusters. Subsequently, we measure the Jaccard similarity index~\cite{jaccard1901etude} between the set of honest clients and the set of all clients that are in a cluster containing at least one honest client -- higher similarity (value closer to $ 1 $) implies that all honest clients are plausibly in one cluster and hence a higher separation between honest and dishonest clients. 

In Figure~\ref{fig:jaccard_similarity} (left), while increasing the flipping probability, Jaccard similarity increases, eventually reaching $ 1 $. Also, by keeping dishonest clients fixed at $ 1 $, the similarity index increases with increasing clients in Figure~\ref{fig:jaccard_similarity} (left). Also, while increasing dishonest clients in Figure~\ref{fig:jaccard_similarity} (right), the similarity decreases, implying hardness of separation with more dishonest clients. \emph{Therefore, Shapley values demonstrate the potential in separating out honest clients.}

\section{\textbf{Conclusion}}
We study the assessment of client contribution in centralized single-server federated learning using Shapley values. We propose  {\framework} to assess client contribution when a subset of clients participate in each training epoch. We discuss a scheduling procedure satisfying two-sided fairness between server and clients for a faster contribution assessment. Empirically, {\framework} is the most efficient method while being consistently accurate in estimating contribution across multiple utility functions. In addition, historic client contributions enable us to analyzing dishonest clients. In future, we extend {\framework} to assess client contribution in a discentralized federated setting. Our framework opens up further research on client selection and federated aggregation based on historic client contributions.

\section*{Acknowledgments}
This Research is supported by the RIE$ 2025 $ Industry Alignment Fund -- Industry Collaboration Project (IAF-ICP) (Award No: I$ 2301 $E$ 0020 $) and Japan-Singapore Joint Call: Japan Science and Technology Agency (JST) and Agency for Science, Technology and Research (A*STAR) $ 2024 $ (Award No: R$ 24 $I$ 6 $IR$ 141 $), administered by A*STAR. D.\ Basu acknowledges the ANR JCJC project REPUBLIC (ANR-$ 22 $-CE$ 23 $-$ 0003 $-$ 01 $), and the PEPR project FOUNDRY (ANR$ 23 $-PEIA-$ 0003 $) for partially supporting this work.

\bibliography{main}

\appendix

\section{Additional Discussion and Proofs}

\begin{replemma}{lm:utility_decomposition}
	\it 
	The total utility in a multi-epoch FL training is the sum of incremental utilities in all the training epochs and the utility of the initial FL model.
	\begin{align*}
		\eval(\flmodel^{(\epoch)}) = \sum_{\epoch = 1}^{\numepochs} \increval(\flmodel^{(\epoch)}, \flmodel^{(\epoch-1)}) + \eval(\flmodel^{(0)}),
	\end{align*}
    where $\eval(\flmodel^{(0)})$ denotes the utility of the initial model.
\end{replemma}

\begin{proof}
	The proof directly follows the decomposition of the total utility $ \eval(\flmodel^{(\epoch)}) $ as the sum of incremental utilities between epoch $ 1 $ to $ \numepochs $ and the utility of the initial model $ \eval(\flmodel^{(0)}) $.
\end{proof}

\begin{replemma}{lm:shapley_value_decomposition}
	\it
	For a client $ \client $, let $ \shapleyvalue_{\client}(\eval) $ be the Shapley value after $ \numepochs $ epochs, and let $ \shapleyvalue^{(\epoch)}_{\client}(\increval) $ be the Shapley value on the incremental utility at epoch $ \epoch $. Since $ \eval $ is a linear sum of $ \increval $, we apply the linearity of Shapley value: the Shapley value w.r.t.\ $ \eval $ is the sum of Shapley values w.r.t.\ $ \increval $ between epoch $1$ to $\numepochs$ and the Shapley value of the initial model. 
	\begin{align*}
		\shapleyvalue_{\client}(\eval)  = \sum_{\epoch=1}^{\numepochs} \shapleyvalue^{(\epoch)}_{\client}(\increval) + \shapleyvalue^{(0)}_{\client}(\eval)
	\end{align*}
\end{replemma}

\begin{proof}
	We prove this lemma by applying linearity property of Shapley value to  Lemma~\ref{lm:utility_decomposition}. 
\end{proof}

\begin{replemma}{lm:non_selected_submodel}
	\it
	Let $ \flmodel^{(\epoch)}_{\subsetofclients} $ be a sub-model consisting of a subset of clients $ \subsetofclients \subseteq \clients $ and $ \client \notin \clients^{(\epoch)} $ be a non-participating client in epoch $ \epoch $. Non-parcipating clients do not influence sub-model reconstruction, formally
	$ \flmodel^{(\epoch)}_{\subsetofclients \cup \{\client\}} = \flmodel^{(\epoch)}_{\subsetofclients} $.
\end{replemma}

\begin{proof}
	Since  $ \client \notin \clients^{(\epoch)} $, we have $ \client \notin \subsetofclients \cap \clients^{(\epoch)}  \Rightarrow \lambda(\client, \subsetofclients \cap \clients^{(\epoch)}) = 0 $. Applying $ \lambda(\client, \subsetofclients \cap \clients^{(\epoch)}) = 0 $ to Eq.~\eqref{eq:fl_sub_model}, 
	{
		\small
		\begin{align*}
			\flmodel^{(\epoch)}_{\subsetofclients \cup \{\client\}} 
		&= \flmodel^{(\epoch-1)} + \sum_{\client' \in \subsetofclients \cup \{\client\}} \lambda(\client', (\subsetofclients \cup \{\client\}) \cap \clients^{(\epoch)})\Delta^{(\epoch)}_{\client}\\
		&= \flmodel^{(\epoch-1)} + \sum_{\client' \in \subsetofclients \cup \{\client\}} \lambda(\client', (\subsetofclients \cap \clients^{(\epoch)}) \cup  (\{\client\} \cap \clients^{(\epoch)}))\Delta^{(\epoch)}_{\client}\\
		&= \flmodel^{(\epoch-1)} + \sum_{\client' \in \subsetofclients \cup \{\client\}} \lambda(\client', (\subsetofclients \cap \clients^{(\epoch)}) \cup  \emptyset)\Delta^{(\epoch)}_{\client}\\
		&= \flmodel^{(\epoch-1)} + \sum_{\client' \in \subsetofclients \cup \{\client\}} \lambda(\client', \subsetofclients \cap \clients^{(\epoch)})\Delta^{(\epoch)}_{\client}\\
		&= \flmodel^{(\epoch-1)} + \lambda(\client, \subsetofclients \cap \clients^{(\epoch)})\Delta^{(\epoch)}_{\client} + \sum_{\client' \in \subsetofclients} \lambda(\client', \subsetofclients \cap \clients^{(\epoch)})\Delta^{(\epoch)}_{\client} \\
		&= \flmodel^{(\epoch-1)} + \sum_{\client' \in \subsetofclients} \lambda(\client', \subsetofclients \cap \clients^{(\epoch)})\Delta^{(\epoch)}_{\client} \\
		&= \flmodel^{(\epoch)}_{\subsetofclients}
		\end{align*}	
	}
\end{proof}

\begin{replemma}{lm:shapley_value_non_selected}
	\it
	At epoch $ t $, the Shapley value of a non-participating client $\client \notin \clients^{(\epoch)} $ with respect to the incremental utility $ \increval $ is zero, $ \shapleyvalue^{(t)}_{\client}(\increval) = 0 $. Therefore, the non-participating client is a null client.
\end{replemma}

\begin{proof}
	According to Lemma~\ref{lm:non_selected_submodel},
	for a non-participating client $ \client $ and a subset of clients $ \subsetofclients \subseteq \clients \setminus \{\client\} $, 
	\begin{align*}
		&\flmodel^{(\epoch)}_{\subsetofclients \cup \{\client\}} = \flmodel^{(\epoch)}_{\subsetofclients}\\
		\Rightarrow & \increval(\flmodel^{(\epoch)}_{\subsetofclients \cup \{\client\}}, \flmodel^{(\epoch-1)}) = \increval(\flmodel^{(\epoch)}_{\subsetofclients}, \flmodel^{(\epoch-1)})
	\end{align*}
	Therefore, the marginal utility of  client $ \client $ is $ 0 $ for each $ \subsetofclients $, and hence, the Shapley value of a non-participating client is $ 0 $ at epoch $ \epoch $.
\end{proof}

\begin{replemma}{lm:shapley_value_complexity}
	\it
	Let $ \numclients $ be the total number of clients and $ \frac{1}{\tau} \in [0,1] $ be the ratio of participating to non-participating clients. In an epoch, the runtime complexity of exactly computing Shapley value is $ \mathcal{O}(2^{\frac{\numclients}{\tau}} + (1 - \frac{1}{\tau})\numclients) $. In $ \numepochs $ epochs, the total running time is $ \mathcal{O}(2^{\frac{\numclients}{\tau}}\numepochs + (1 - \frac{1}{\tau})\numclients\numepochs) $.
\end{replemma}

\begin{proof} 
	In the direct approach for an exact Shapley value computation of $ \numclients $ players, $ 2^{\numclients} $ unique subsets of players are enumerated, resulting in $ \mathcal{O}(2^{\numclients}) $ running time.
	
	In the context of FL at training epoch $ \epoch $, the number of non-participating clients is $ (1 - \frac{1}{\tau})\numclients $. Their Shapley value is deterministically computed as $ 0 $ in $ (1 - \frac{1}{\tau})\numclients $ running time. 
	
	For the  $  \frac{\numclients}{\tau} $ participating clients, we need to enumerate $ 2^{\frac{\numclients}{\tau}} $ unique subsets of clients, resulting in $ \mathcal{O}(2^{\frac{\numclients}{\tau}}) $ running time for the selected clients. Thus, the total running time is $ \mathcal{O}(2^{\frac{\numclients}{\tau}} + (1 - \frac{1}{\tau})\numclients)  $. 
	
	If $ \tau $ is fixed in each epoch, by repeating the same analysis, the total running time in $ \numepochs $ epochs is $ \mathcal{O}(2^{\frac{\numclients}{\tau}}\numepochs + (1 - \frac{1}{\tau})\numclients\numepochs) $. 
\end{proof}

\begin{replemma}{lm:error_bound}
	\it
	If we apply an $\epsilon$-approximation algorithm to compute the incremental Shapley values at each epoch, the total estimation error in the global Shapley value is $\mathcal{O}(\frac{\numepochs\epsilon}{\tau})$, which is of the same order as existing algorithms. 
\end{replemma}

\begin{proof}
	Since $ \numepochs $ is the total epochs and $ \tau $ is the ratio of participating to non-participating clients, a client participates in $ \frac{\numepochs}{\tau} $ epochs in expectation. Therefore, The Shapley value of participating clients is approximated $ \frac{\numepochs}{\tau} $ times; in rest of the  $ \numepochs - \frac{\numepochs}{\tau} $ epochs, the Shapley value of non-participating clients is exactly $ 0 $.

	Since, the approximation error is $ \epsilon $ in each epoch, total estimation error in the global Shapley value is $ \mathcal{O}(\frac{\numepochs\epsilon}{\tau}) $.

\end{proof}

\begin{replemma}{lm:complexity_scheduling}
	\it
	For $ \maxepochs \le \numepochs $ denoting the maximum number of epochs for Shapley value computation, the runtime complexity is $ \mathcal{O}(2^{\frac{\numclients}{\tau}}\maxepochs + (1 - \frac{1}{\tau})\numclients\maxepochs) $, which is $\frac{\maxepochs}{\numepochs}$ fraction of the total runtime complexity without scheduling.
\end{replemma}

\begin{proof}
	The proof follows the proof of Lemma~\ref{lm:shapley_value_complexity} by replacing $ \numepochs $ with $ \maxepochs $.
\end{proof}

\subsection{Derivation of the Lower Bound of Two-sided Fairness. }
\begin{align*}
	&\sum_{\epoch = 1}^{\numepochs} \varepochweight^{(\epoch)} \varepoch^{(\epoch)}  -  \gamma \sum_{\client, \client' \in \clients} |\sum_{\epoch=1}^{\numepochs} \varclientpos^{(\epoch)}_{\client} \varepoch^{(\epoch)} - \sum_{\epoch=1}^{\numepochs} \varclientpos^{(\epoch)}_{\client'} \varepoch^{(\epoch)}|\\
	=&\sum_{\epoch = 1}^{\numepochs} \varepochweight^{(\epoch)} \varepoch^{(\epoch)}  -  \gamma \sum_{\client, \client' \in \clients} |\sum_{\epoch=1}^{\numepochs} (\varclientpos^{(\epoch)}_{\client} -  \varclientpos^{(\epoch)}_{\client'}) \varepoch^{(\epoch)}|\\
	\ge&\sum_{\epoch = 1}^{\numepochs} \varepochweight^{(\epoch)} \varepoch^{(\epoch)}  -  \gamma \sum_{\client, \client' \in \clients} \sum_{\epoch=1}^{\numepochs} | (\varclientpos^{(\epoch)}_{\client} -  \varclientpos^{(\epoch)}_{\client'}) \varepoch^{(\epoch)}|\\
	=&\sum_{\epoch = 1}^{\numepochs} \varepochweight^{(\epoch)} \varepoch^{(\epoch)}  -  \gamma \sum_{\client, \client' \in \clients} \sum_{\epoch=1}^{\numepochs} | (\varclientpos^{(\epoch)}_{\client} -  \varclientpos^{(\epoch)}_{\client'})|\varepoch^{(\epoch)}\\
	=&\sum_{\epoch = 1}^{\numepochs} \varepochweight^{(\epoch)} \varepoch^{(\epoch)}  -  \gamma  \sum_{\epoch=1}^{\numepochs} \sum_{\client, \client' \in \clients} | (\varclientpos^{(\epoch)}_{\client} -  \varclientpos^{(\epoch)}_{\client'})|\varepoch^{(\epoch)}\\
	=&\sum_{\epoch = 1}^{\numepochs} \varepochweight^{(\epoch)} \varepoch^{(\epoch)}  -  \gamma  \sum_{\epoch=1}^{\numepochs} \varepoch^{(\epoch)} \sum_{\client, \client' \in \clients} | (\varclientpos^{(\epoch)}_{\client} -  \varclientpos^{(\epoch)}_{\client'})|\\
	=&\sum_{\epoch = 1}^{\numepochs} (\varepochweight^{(\epoch)}   -  \gamma  \sum_{\client, \client' \in \clients} | (\varclientpos^{(\epoch)}_{\client} -  \varclientpos^{(\epoch)}_{\client'})|) \varepoch^{(\epoch)}
\end{align*}

\subsection{Greedy Model Aggregation}
In FL, the server aggregates the local gradients of all selected clients $ \clients^{(\epoch)} $ to construct a global model according to Eq.~\eqref{eq:fl_naive}. However, considering all local gradients may not achieve best prediction performance. In this context, Shapley value is effective to choose the optimal subset of clients for the global model. Let $ \subsetofclients^* \subseteq \clients^{(\epoch)} $ be the optimal subset based on utility on the validation dataset. When the server aims to \emph{minimize the utility function} (such as training loss) we compute $ \subsetofclients^* $ as the subset of clients achieving the \emph{lowest incremental utility}. 

\begin{align}
	& \subsetofclients^* = \argmin_{\subsetofclients \subseteq \clients^{(\epoch)}} \increval(\flmodel^{(\epoch)}_{\subsetofclients}, \flmodel^{(\epoch-1)})\\
	&\flmodel^{(\epoch)}
	= \flmodel^{(\epoch-1)} + \sum_{\client \in \subsetofclients^*} \lambda(\client, \subsetofclients^* \cap \clients^{(\epoch)})\Delta^{(\epoch)}_{\client}
\end{align}	

\begin{figure*}[!t]
	\centering
	\includegraphics[scale=0.4]{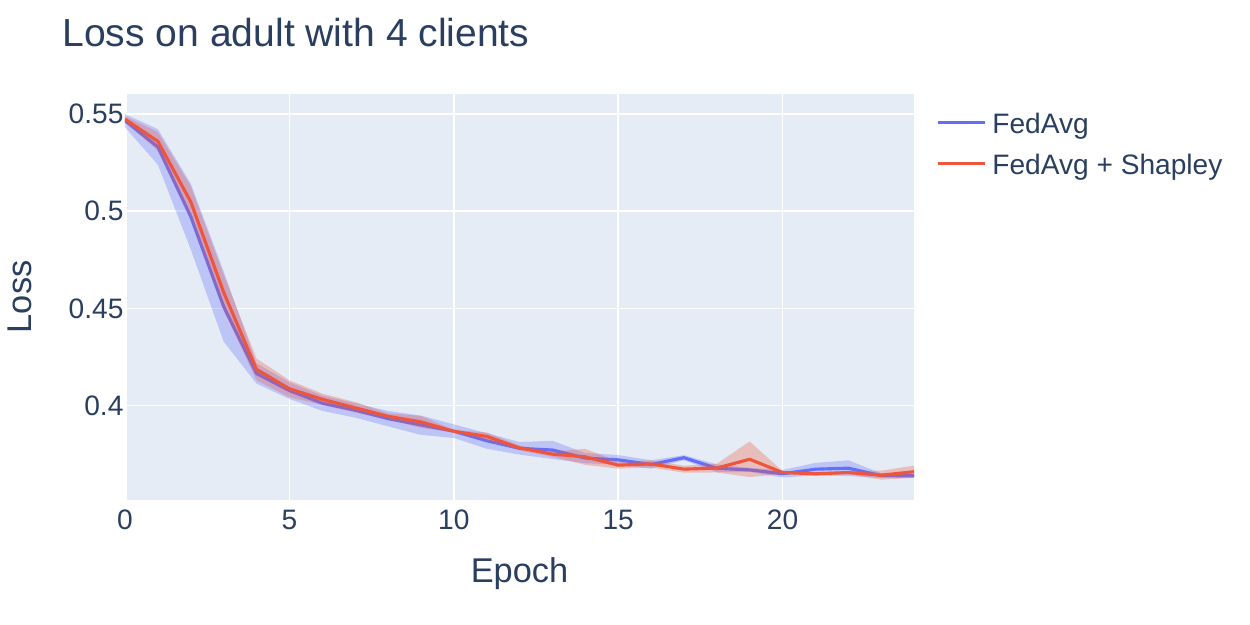}
	\includegraphics[scale=0.4]{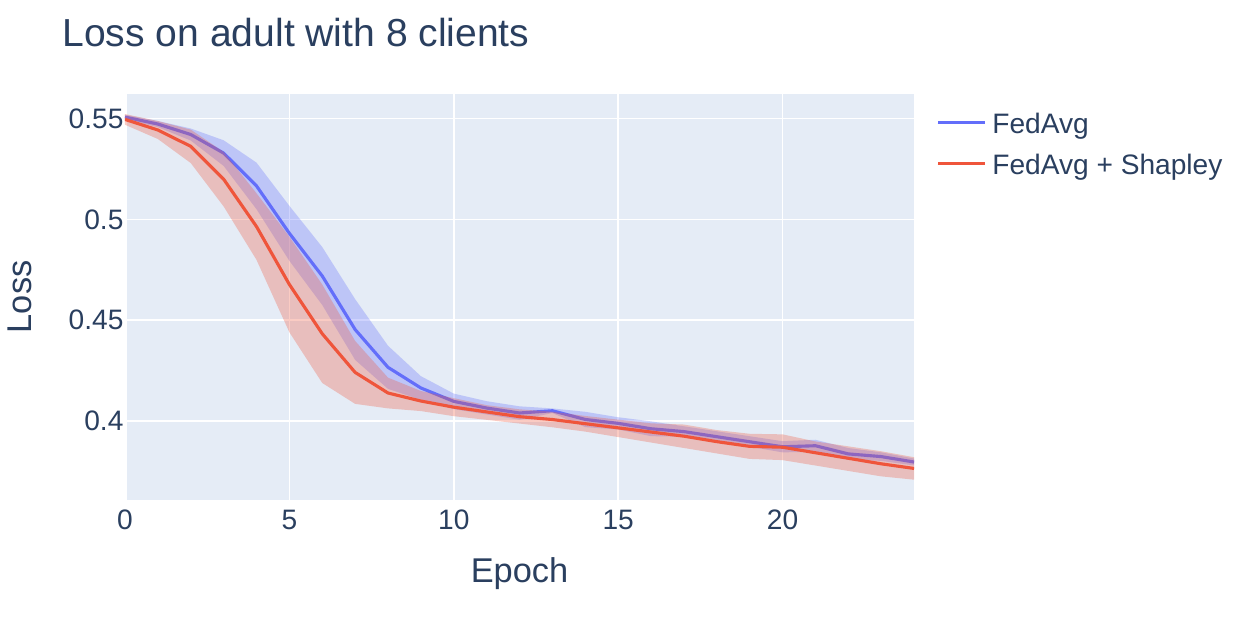}
	\includegraphics[scale=0.4]{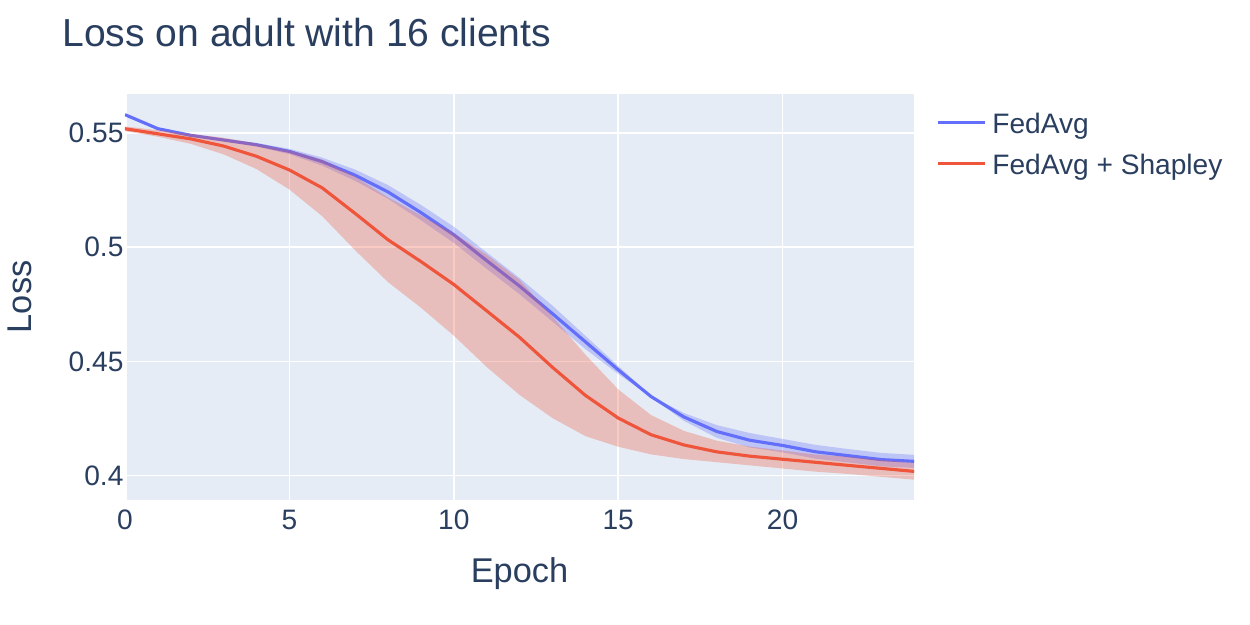}
	\includegraphics[scale=0.4]{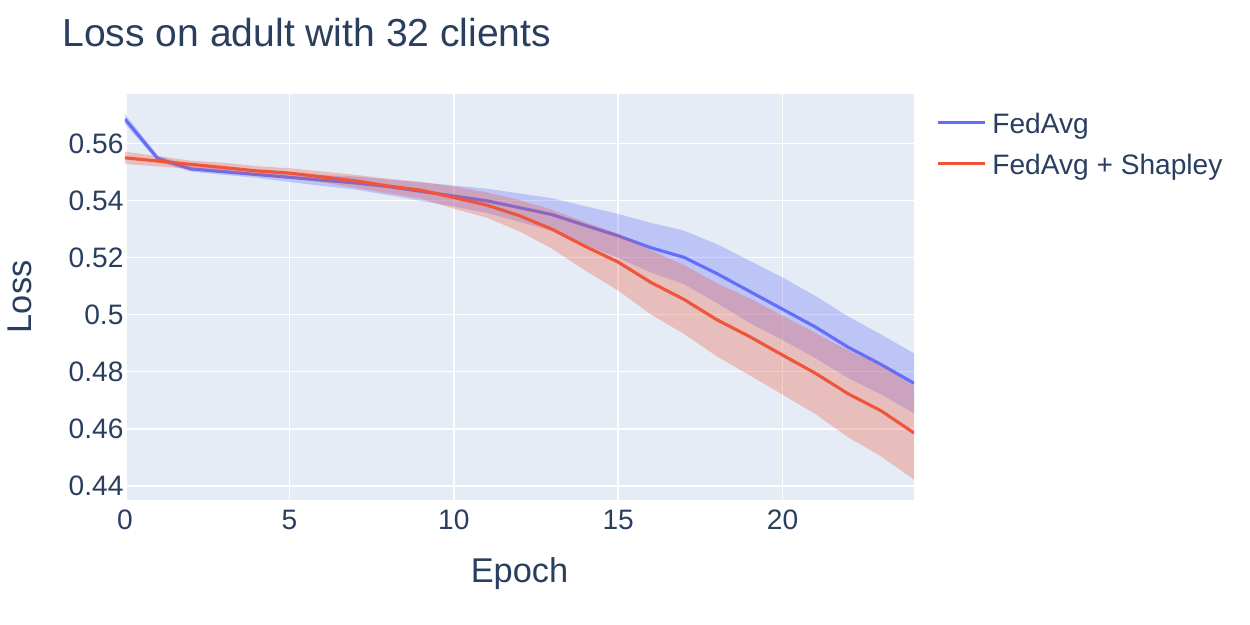}
	
	\caption{In Adult dataset, we demonstrate the impact of training loss vs.\ epochs when FedAvg is combined with Shapley value computed by {\framework} (referred as `FedAvg + Shapley' in the plot). When Shapley value is used to select the optimal subset of local models for federated aggregation, the training loss decreases quickly. }
	\label{fig:adult_loss}
\end{figure*}

\begin{figure*}[!t]
	\centering
	\includegraphics[scale=0.4]{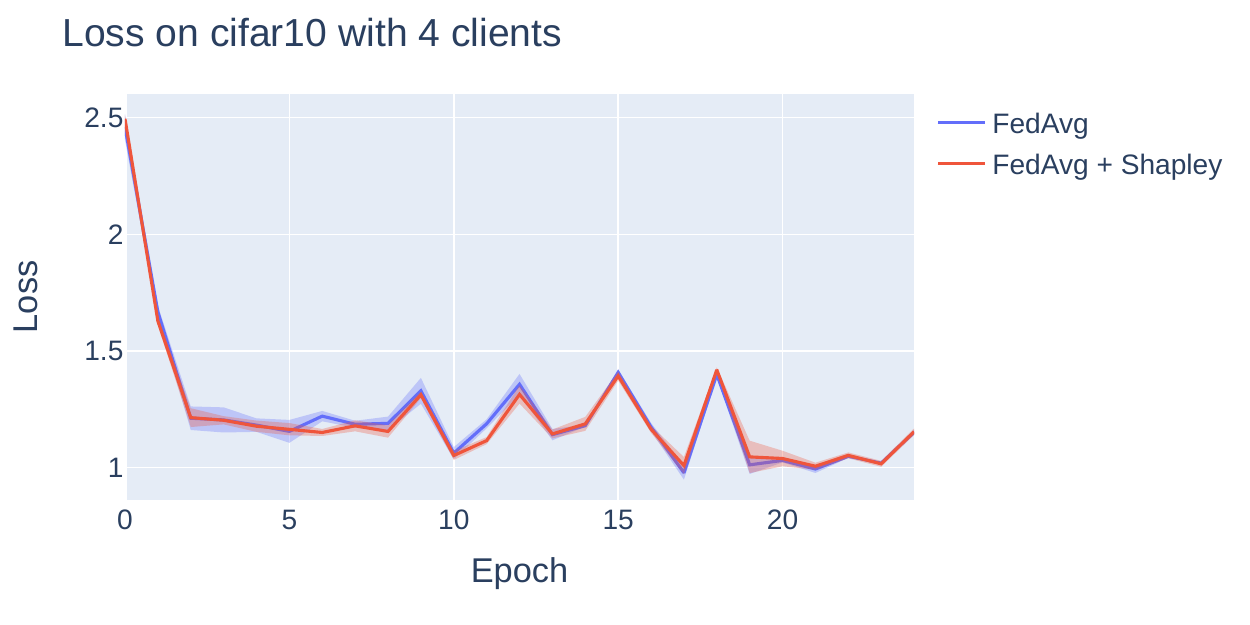}
	\includegraphics[scale=0.4]{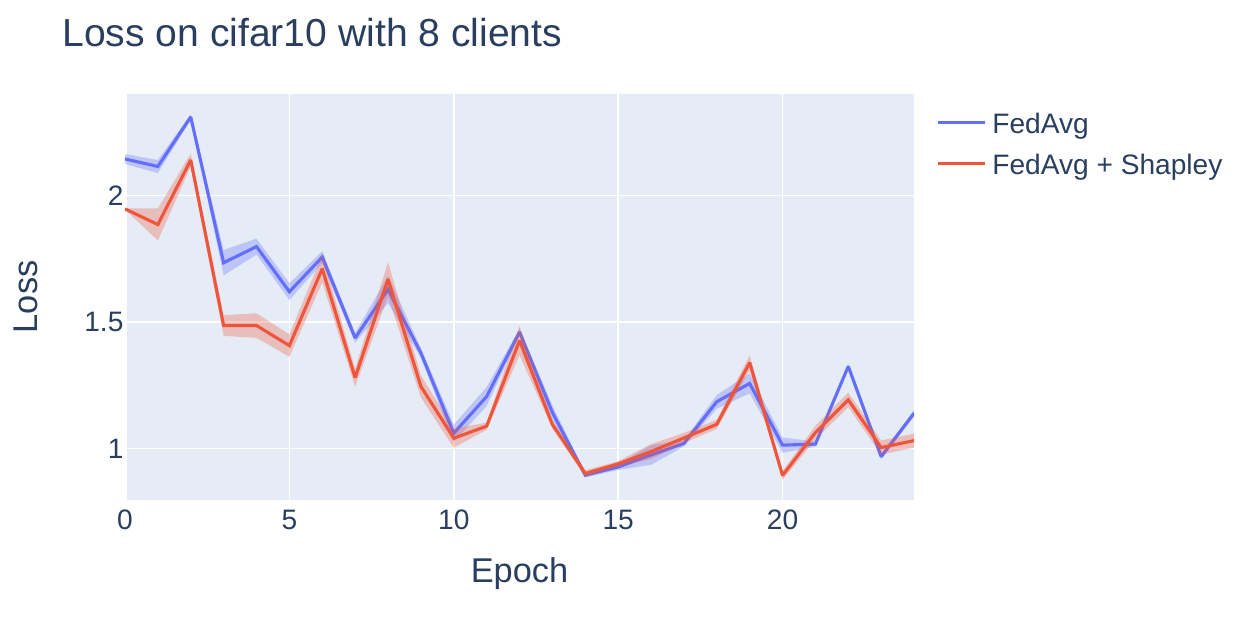}
	\includegraphics[scale=0.4]{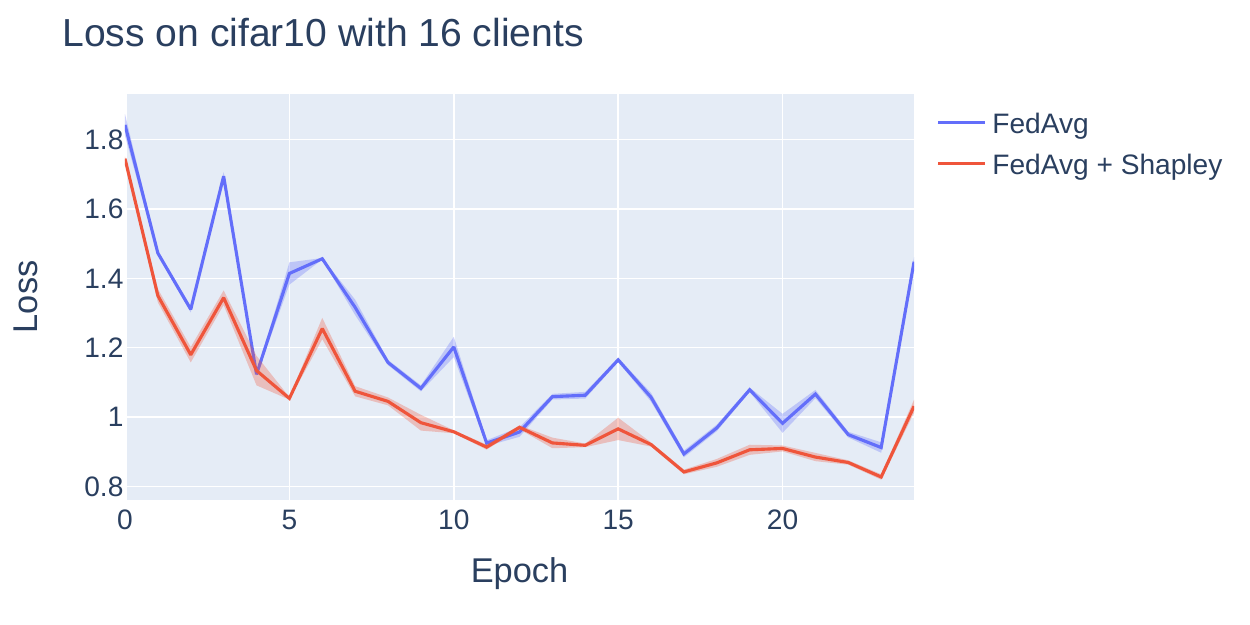}
	\includegraphics[scale=0.4]{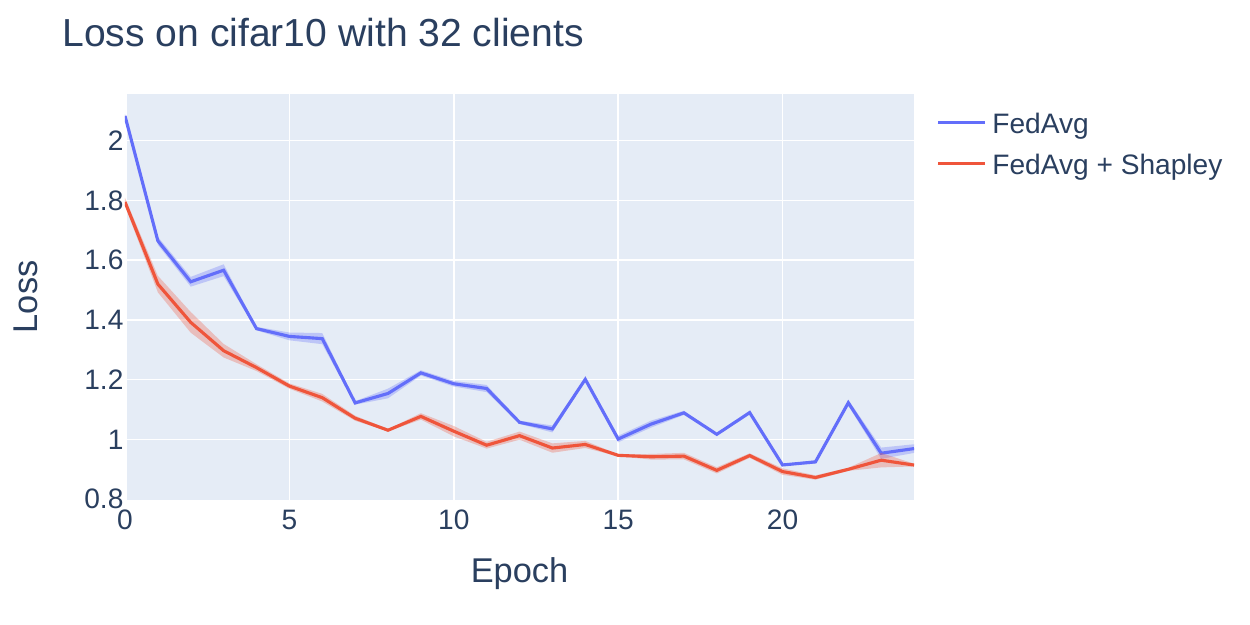}
	
	\caption{In CIFAR10 dataset, we demonstrate the impact of training loss vs.\ epochs when FedAvg is combined with Shapley value computed by {\framework}. When Shapley value is used to select the optimal subset of local models for federated aggregation, the training loss decreases quickly. }
	\label{fig:cifar10_loss}
\end{figure*}

Computing $ \subsetofclients^* $ comes as a byproduct of Shapley value and we leverage $ \subsetofclients^* $ to find the global model. We demonstrate empirical evidences in Figure~\ref{fig:adult_loss} and~\ref{fig:cifar10_loss}.

\section{Experiments Extended}
\label{sec:app_exp}

\paragraph{FL Training Setup.} We run experiments in a cluster containing NVIDIA GeForce RTX $3090$ GPU with $24$ GB GPU memory and 125 GB CPU memory. We consider a non i.i.d.\ data distribution of clients following a Beta distribution with parameter $\beta=50$ for Adult, $0.5$ for COMPAS, and $0.25$ for CIFAR10. Intuitively, a lower value of $\beta$ denotes higher non i.i.d.\ distribution. We train binary classification datasets such as Adult and COMPAS using a MLP model with $4$ fully connected layers with output dimension $[64, 128, 256, 512]$ followed by a classification layer. For image classification on CIFAR10 dataset, we consider a 2D CNN model with $3$ depth, $128$ width with batch normalization.  In FL, we vary the number of clients in $\{4, 8, 16, 32, 64\}$ and the number of epochs in $\{12, 25, 37, 50\}$. In each epoch, $50 \%$ of the clients are uniformly chosen to participate in training.  Each client performs $10$ local epochs of batch training with batch size $64$ and learning rate $0.001$.

\begin{figure*}[!t]
	\centering
	\includegraphics[scale=0.45]{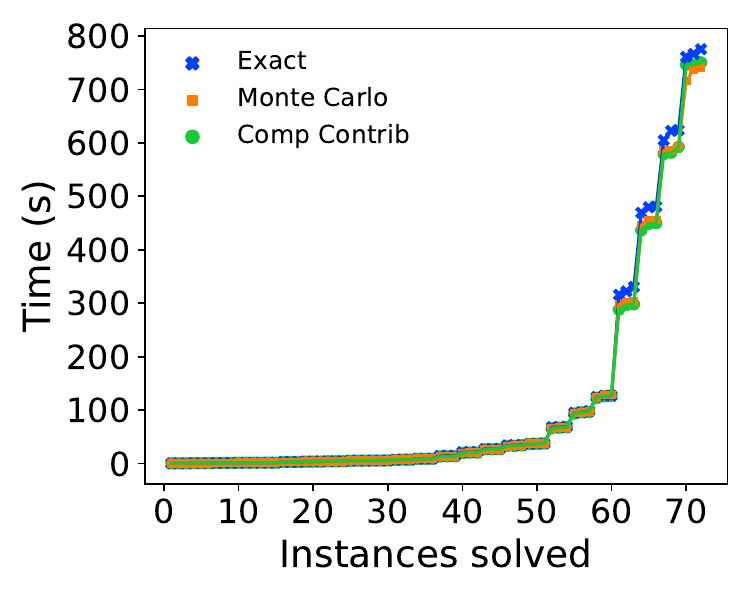}
	\includegraphics[scale=0.45]{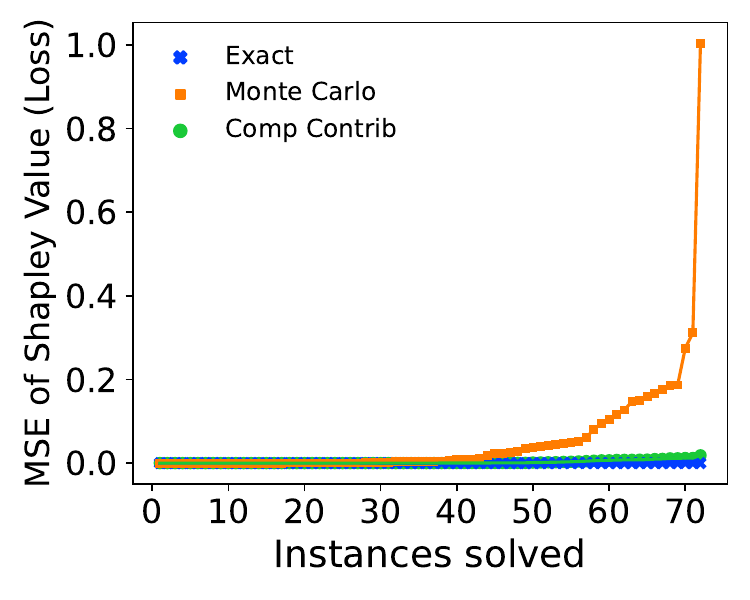}
	\includegraphics[scale=0.45]{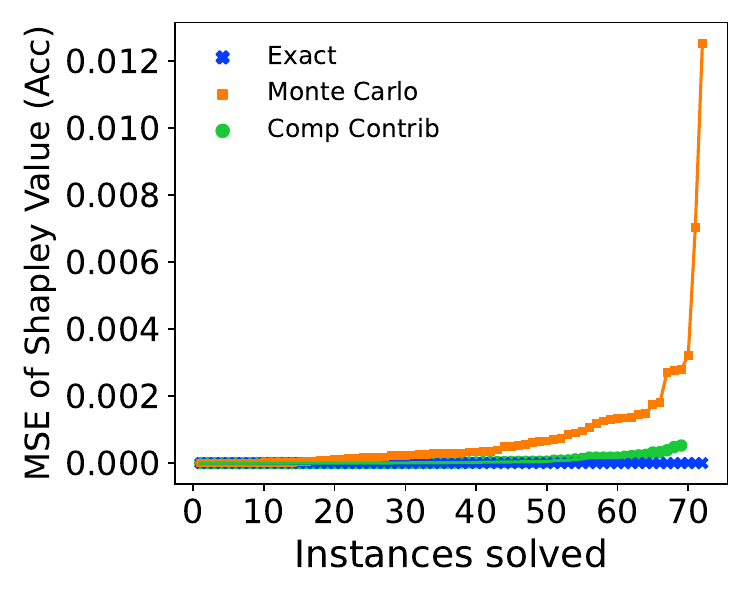}
	\caption{Comparative performance of different approximation algorithms for Shapley value computation. Complementary contribution based method has superior performance than Monte Carlo sampling in both computation time and estimation accuracy of Shapley values.\\~\\To achieve efficiency, complementary contribution based technique for computing Shapley value applies a carefully designed stratified sampling than random sampling of Monte Carlo. For a given error threshold on estimated Shapley values, the first is shown to have a better sample complexity than the latter.}
\end{figure*}

\begin{figure*}
	\centering
	\includegraphics[scale=0.42]{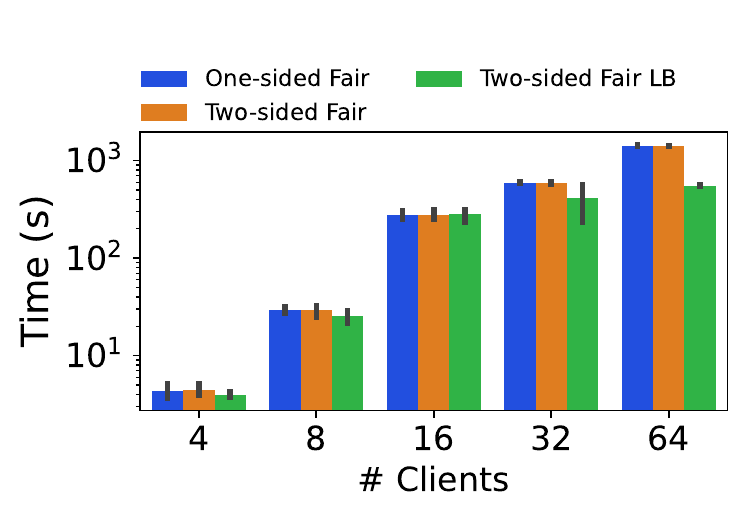}
	\includegraphics[scale=0.42]{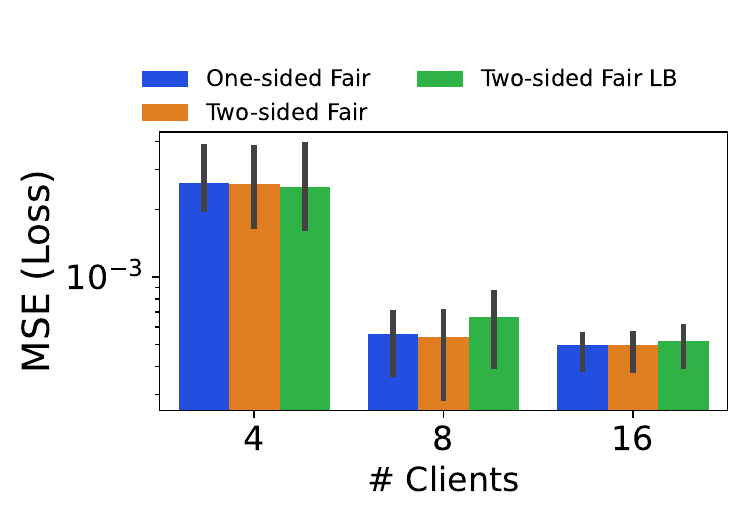}
	\includegraphics[scale=0.42]{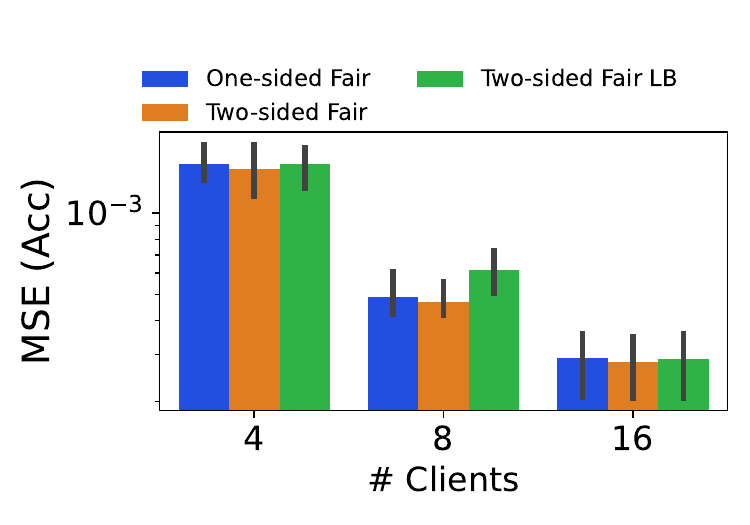}
	\caption{Impact of different scheduling procedures on Shapley value computation time, MSE of Shapley value on model loss and on model accuracy when $ \frac{\maxepochs}{\numepochs} = 0.25 $.}
\end{figure*}

\begin{figure*}
	\centering
	\includegraphics[scale=0.42]{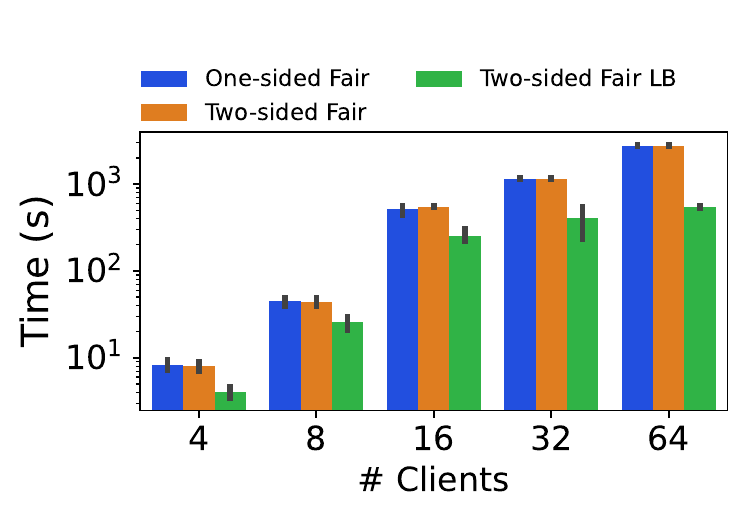}
	\includegraphics[scale=0.42]{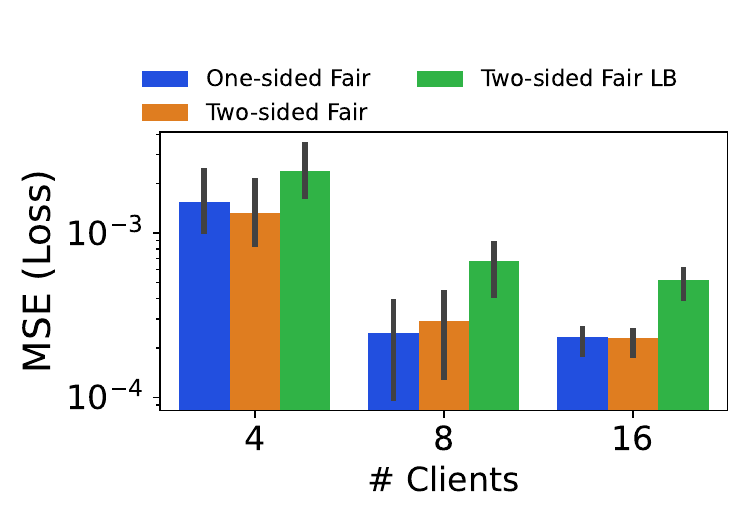}
	\includegraphics[scale=0.42]{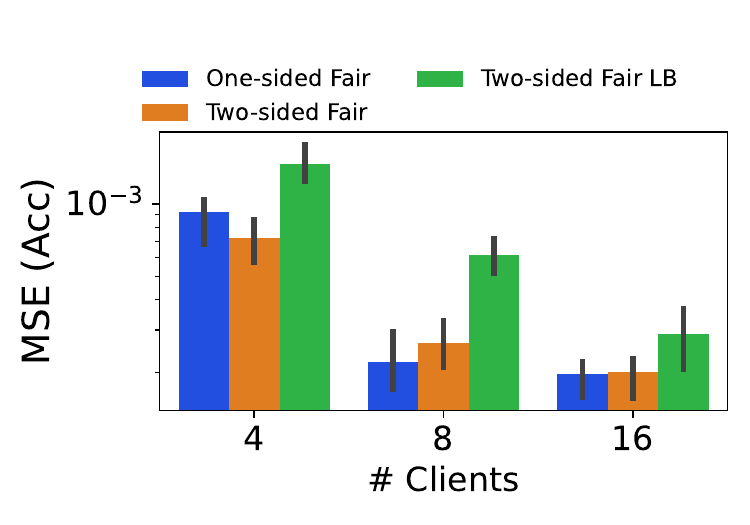}
	\caption{Impact of different scheduling procedures on Shapley value computation time, MSE of Shapley value on model loss and on model accuracy when $ \frac{\maxepochs}{\numepochs} = 0.5 $.}
\end{figure*}

\begin{figure*}
	\centering
	\includegraphics[scale=0.42]{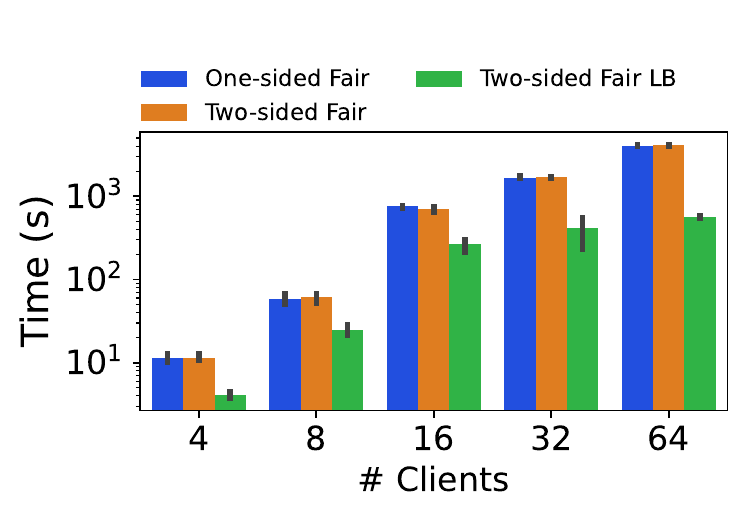}
	\includegraphics[scale=0.42]{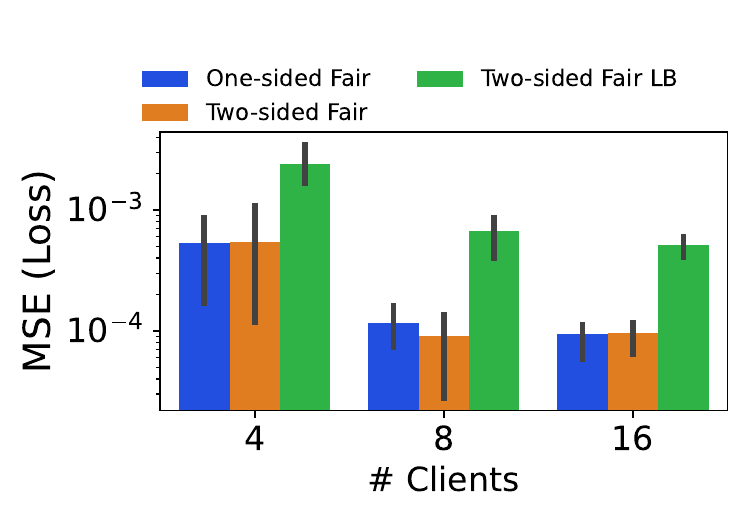}
	\includegraphics[scale=0.42]{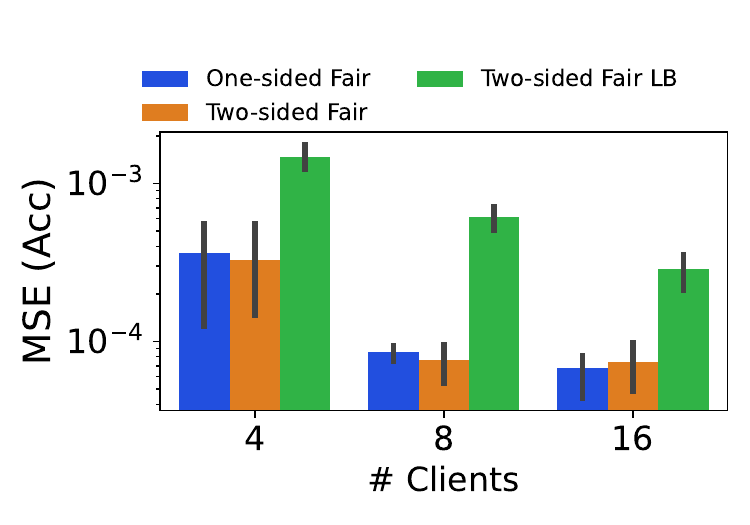}
	\caption{Impact of different scheduling procedures on Shapley value computation time, MSE of Shapley value on model loss and on model accuracy when $ \frac{\maxepochs}{\numepochs} = 0.75 $.}
\end{figure*}

\begin{figure*}
	\centering
	\includegraphics[scale=0.42]{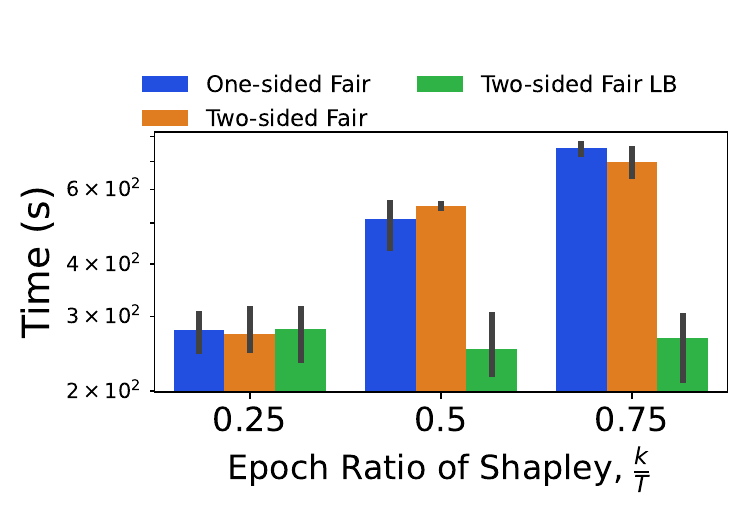}
	\includegraphics[scale=0.42]{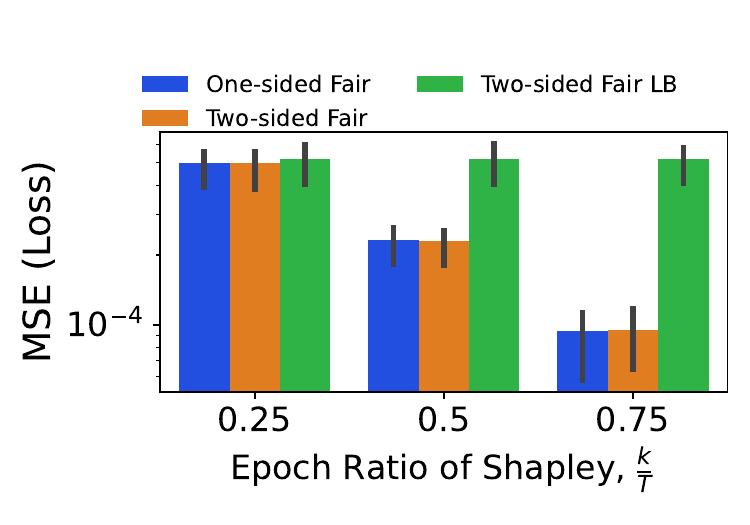}
	\includegraphics[scale=0.42]{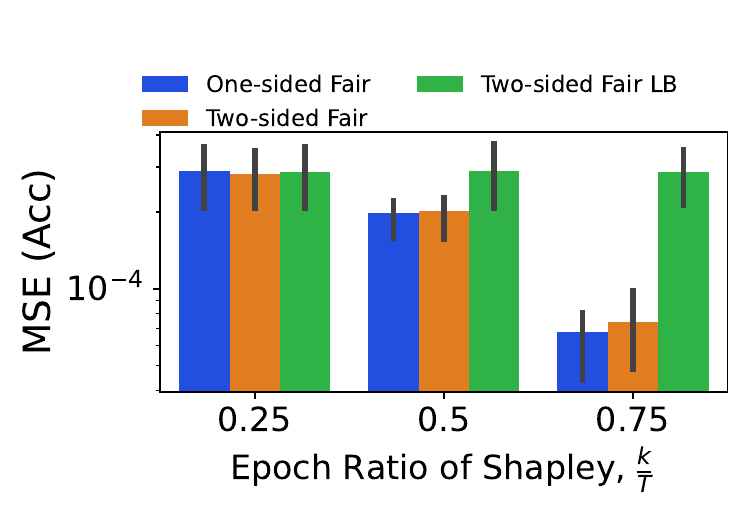}
	\caption{Impact of parameter $ k $ -- the maximum epochs for Shapley computation -- on Shapley value computation time, MSE of Shapley value on model loss and on model accuracy in Adult Dataset.}
\end{figure*}

\begin{figure*}
	\centering
	\includegraphics[scale=0.42]{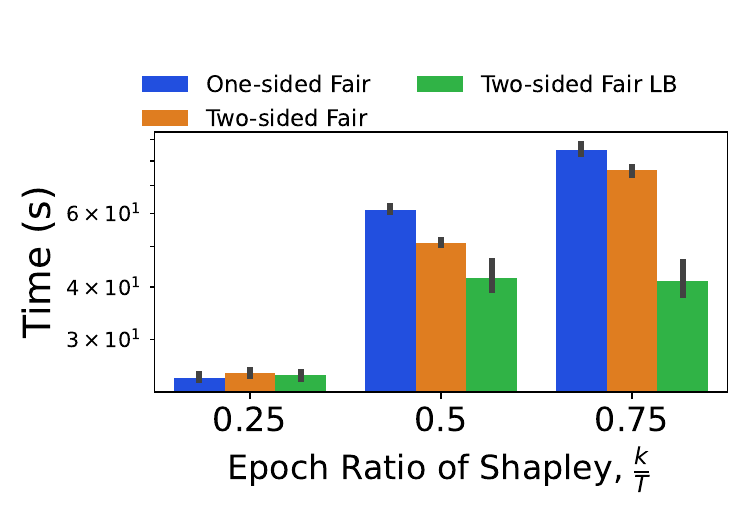}
	\includegraphics[scale=0.42]{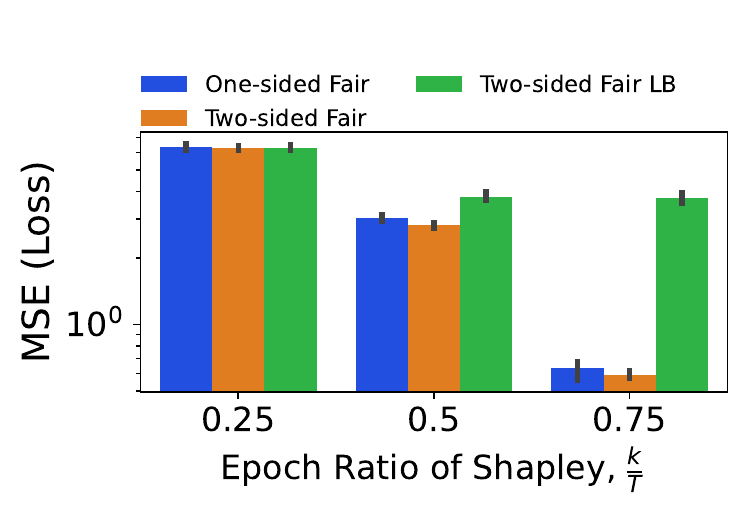}
	\includegraphics[scale=0.42]{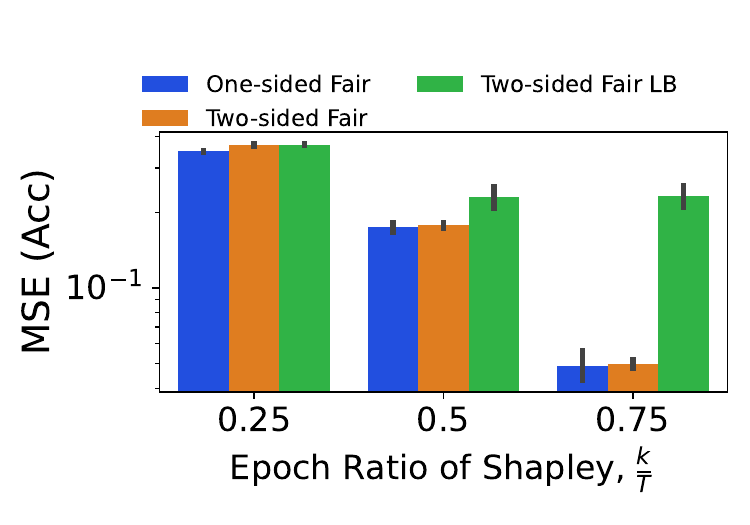}
	\caption{Impact of parameter $ k $ -- the maximum epochs for Shapley computation -- on Shapley value computation time, MSE of Shapley value on model loss and on model accuracy in COMPAS Dataset.}
\end{figure*}

\begin{figure*}
	\centering
	\includegraphics[scale=0.45]{figures/client_intention_analysis/client_intent_analysis_vary_clients_5_0.5_1.pdf}
	\includegraphics[scale=0.45]{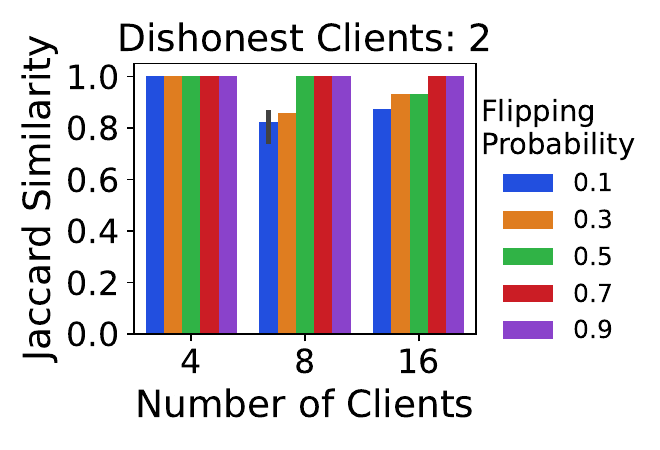}
	\includegraphics[scale=0.45]{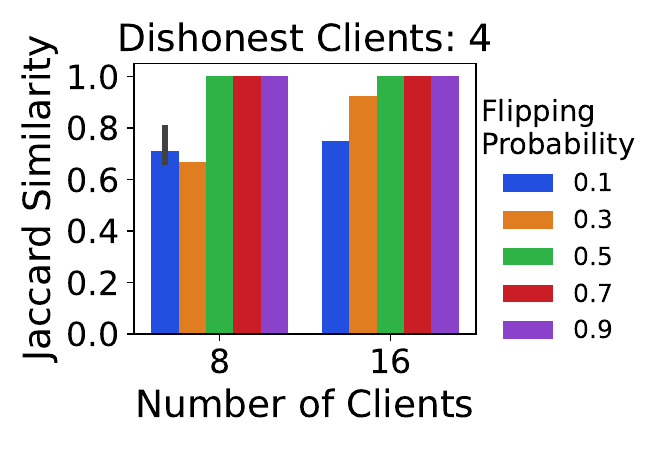}
	\includegraphics[scale=0.45]{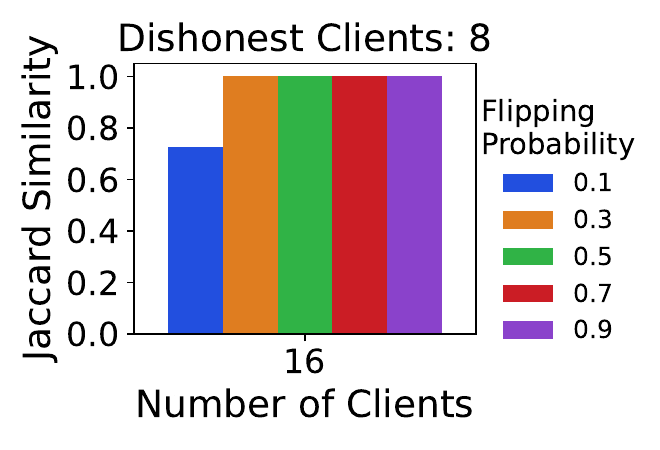}
	\caption{Effectiveness of separating honest clients -- a higher Jaccard similarity denotes a higher separation. We consider  flipping window size $ 5 $. In each epoch, $ 50\% $ clients participate in training.}
\end{figure*}

\begin{figure*}
	\centering
	\includegraphics[scale=0.45]{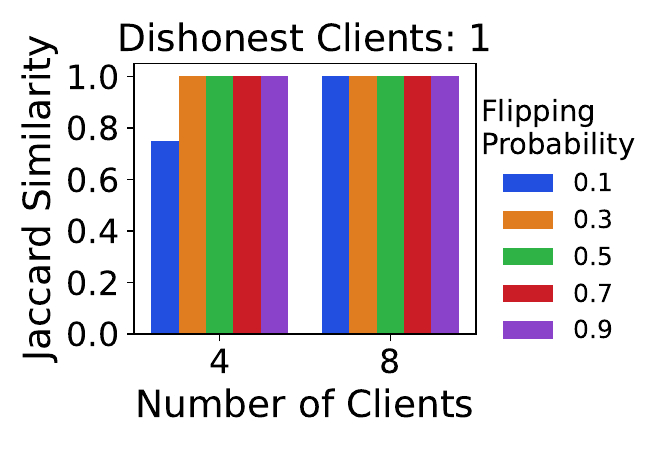}
	\includegraphics[scale=0.45]{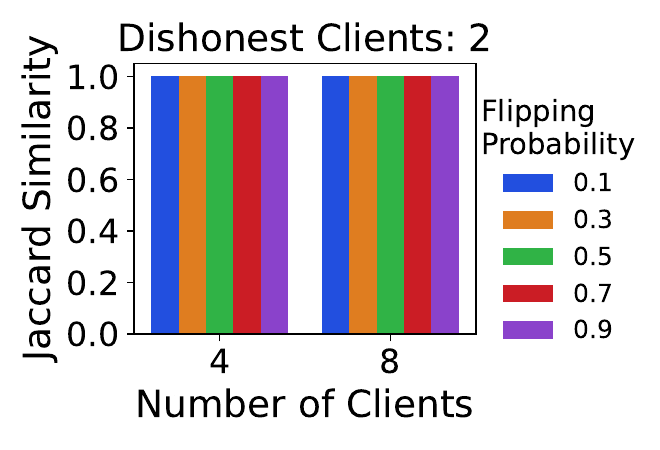}
	\includegraphics[scale=0.45]{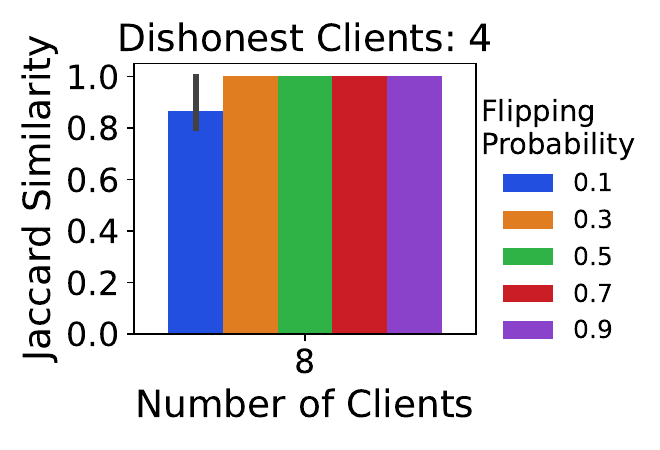}
	\includegraphics[scale=0.45]{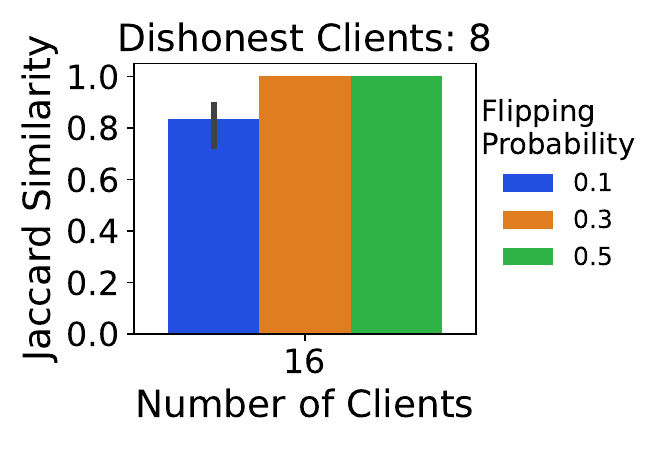}
	\caption{Effectiveness of separating honest clients -- a higher Jaccard similarity denotes a higher separation. We consider  flipping window size $ 5 $. In each epoch, $ 100\% $ clients participate in training.}
\end{figure*}

\begin{figure*}
	\centering
	\includegraphics[scale=0.45]{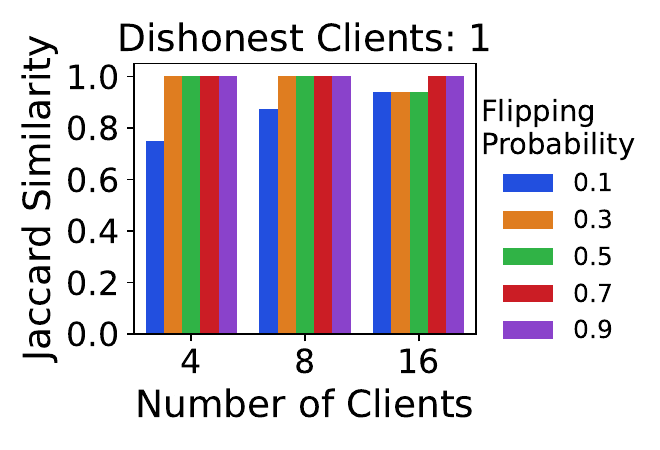}
	\includegraphics[scale=0.45]{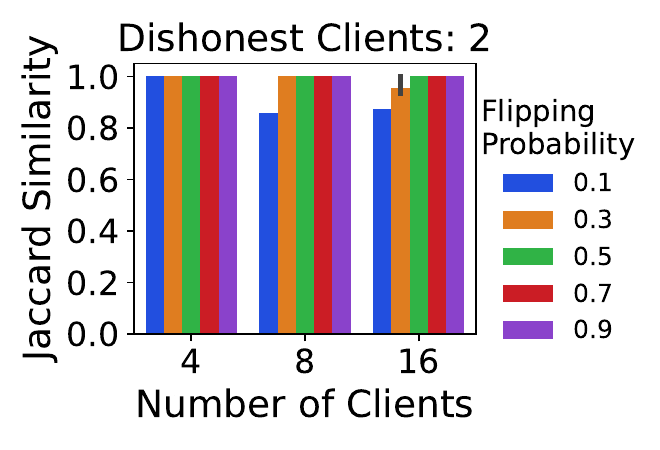}
	\includegraphics[scale=0.45]{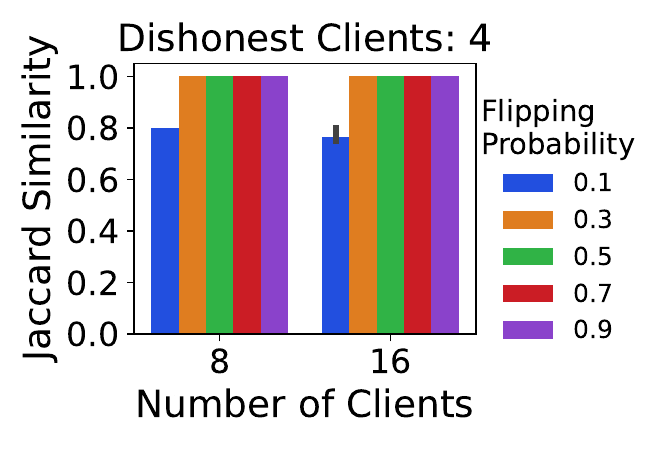}
	\includegraphics[scale=0.45]{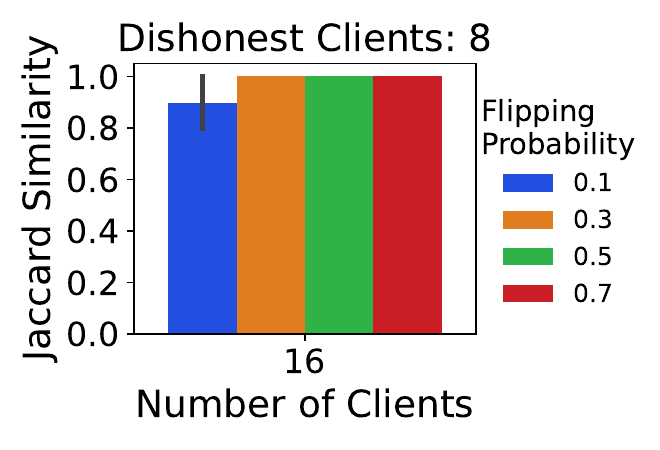}
	\caption{Effectiveness of separating honest clients -- a higher Jaccard similarity denotes a higher separation. We consider  flipping window size $ 10 $. In each epoch, $ 50\% $ clients participate in training.}
\end{figure*}

\begin{figure*}
	\centering
	\includegraphics[scale=0.45]{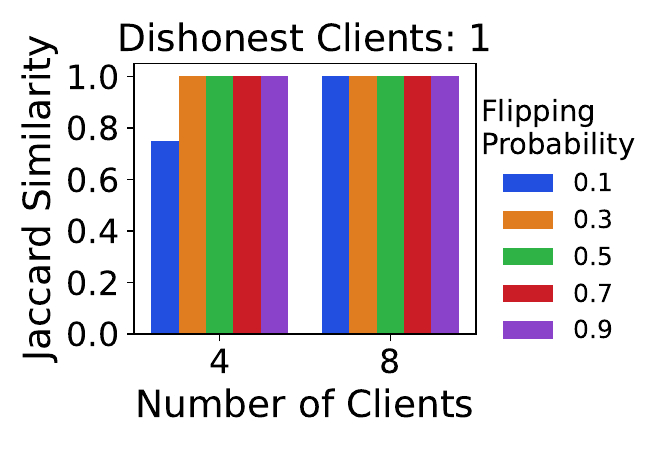}
	\includegraphics[scale=0.45]{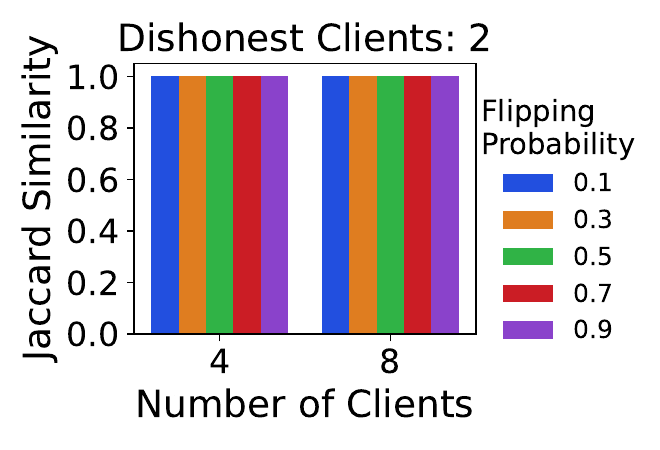}
	\includegraphics[scale=0.45]{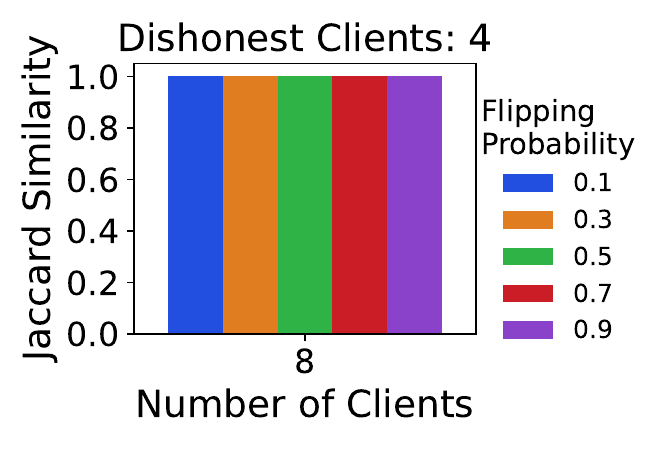}
	\includegraphics[scale=0.45]{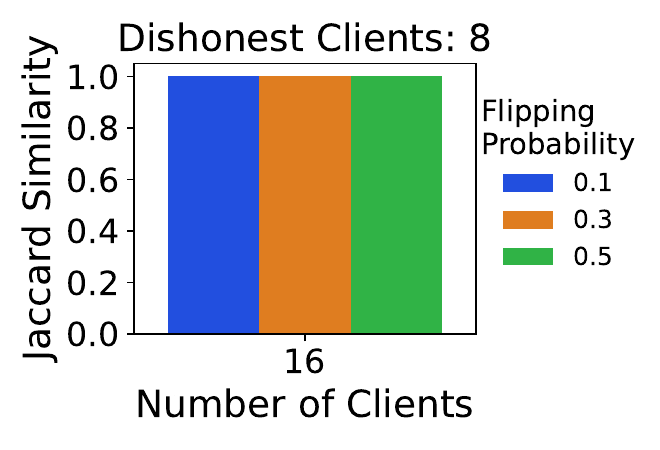}
	\caption{Effectiveness of separating honest clients -- a higher Jaccard similarity denotes a higher separation. We consider  flipping window size $ 10 $. In each epoch, $ 100\% $ clients participate in training.}
\end{figure*}

\begin{figure*}
	\centering
	\includegraphics[scale=0.4]{figures/client_intention_analysis/client_intent_analysis_vary_dishonest_clients_5_0.5_0.1.pdf}
	\includegraphics[scale=0.4]{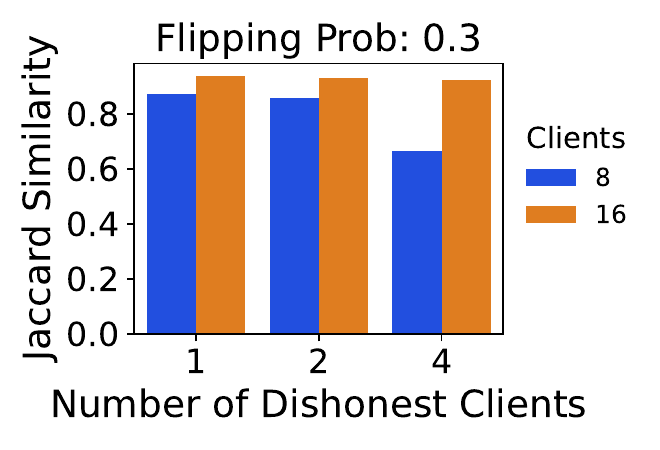}	
	\includegraphics[scale=0.4]{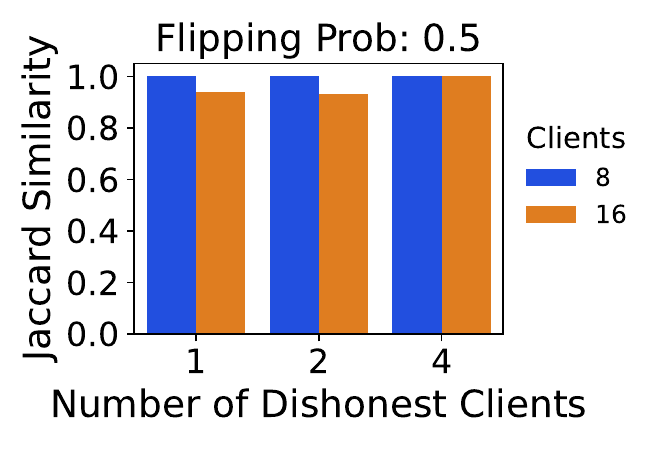}
	\includegraphics[scale=0.4]{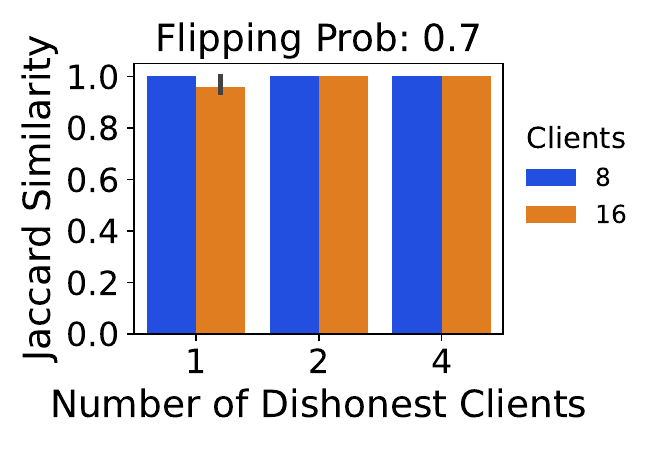}
	\includegraphics[scale=0.4]{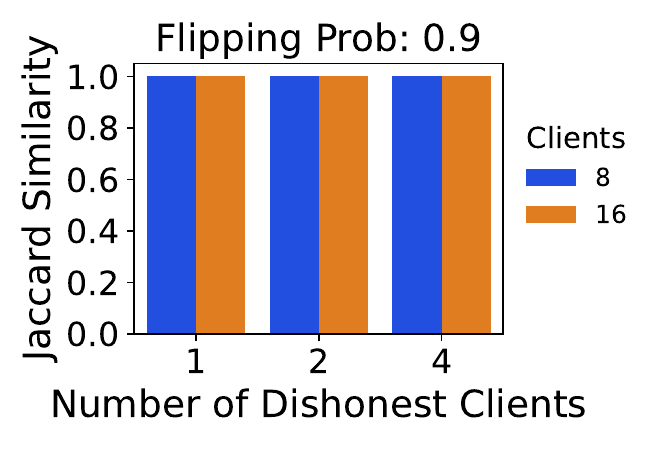}
	\caption{Effectiveness of separating honest clients while varying dishonest clients. A higher Jaccard similarity denotes a higher separation. We consider flipping window size $ 5 $. In each epoch, $ 50\% $ clients participate in training. }
\end{figure*}

\begin{figure*}
	\centering
	\includegraphics[scale=0.4]{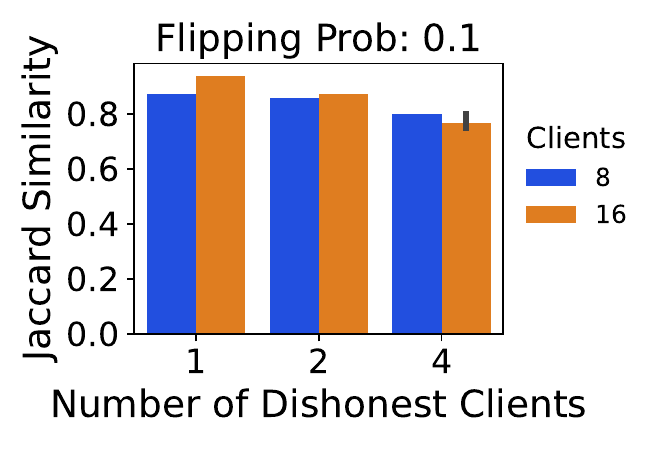}
	\includegraphics[scale=0.4]{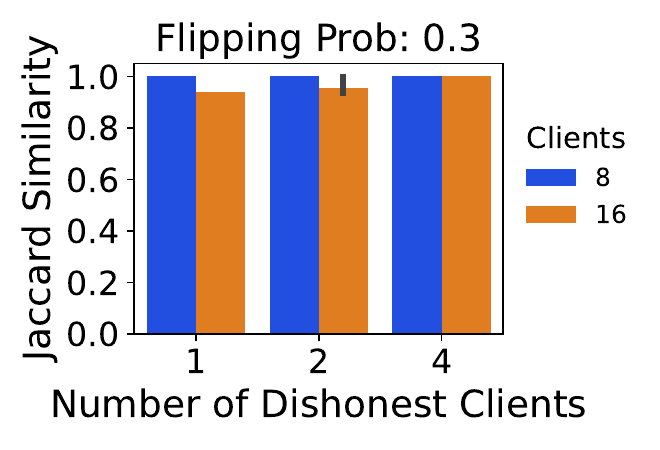}	
	\includegraphics[scale=0.4]{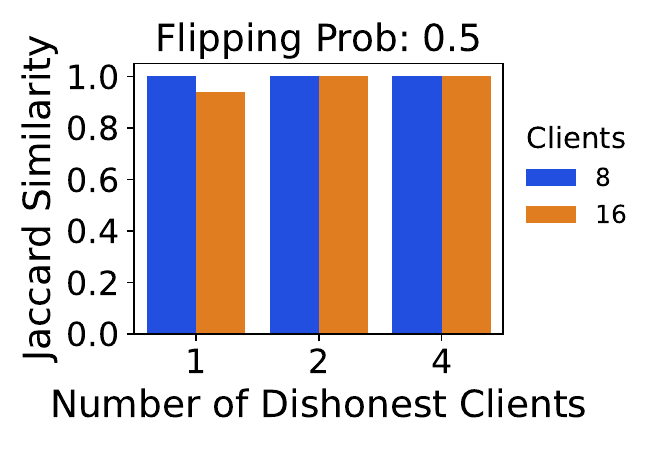}
	\includegraphics[scale=0.4]{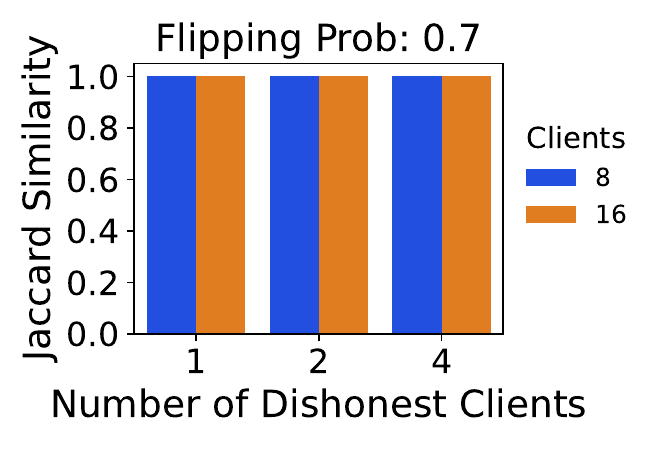}
	\includegraphics[scale=0.4]{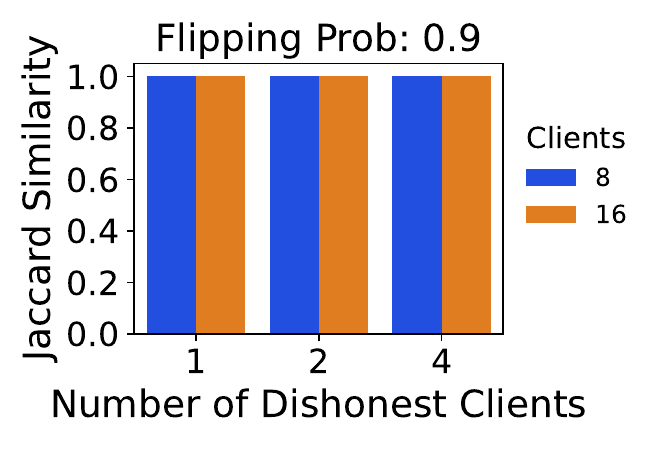}
	\caption{Effectiveness of separating honest clients while varying dishonest clients. A higher Jaccard similarity denotes a higher separation. We consider flipping window size $ 10 $. In each epoch, $ 50\% $ clients participate in training.}
\end{figure*}

\begin{figure*}
	\centering
	\includegraphics[scale=0.4]{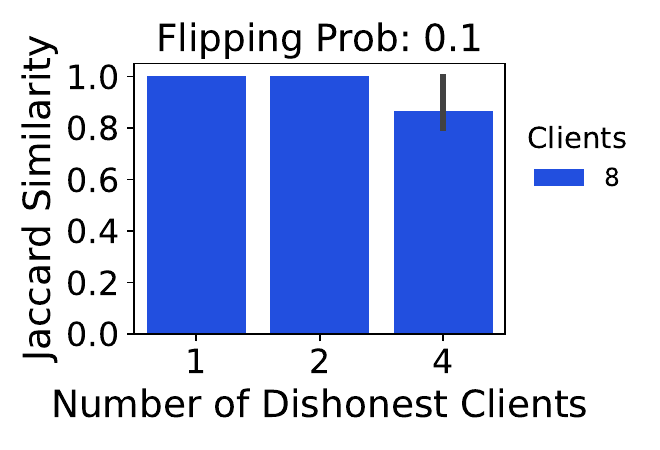}
	\includegraphics[scale=0.4]{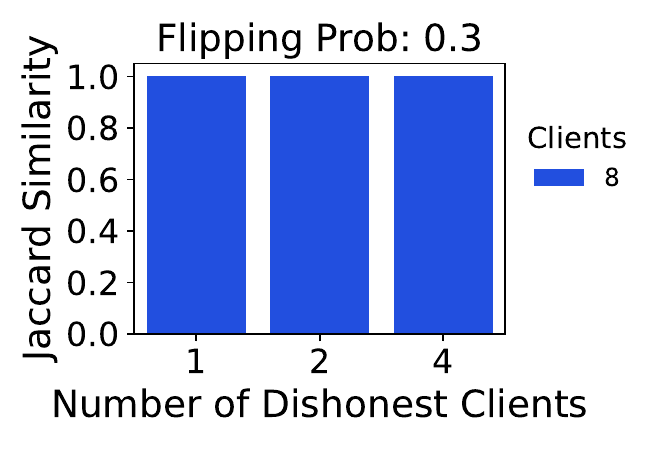}	
	\includegraphics[scale=0.4]{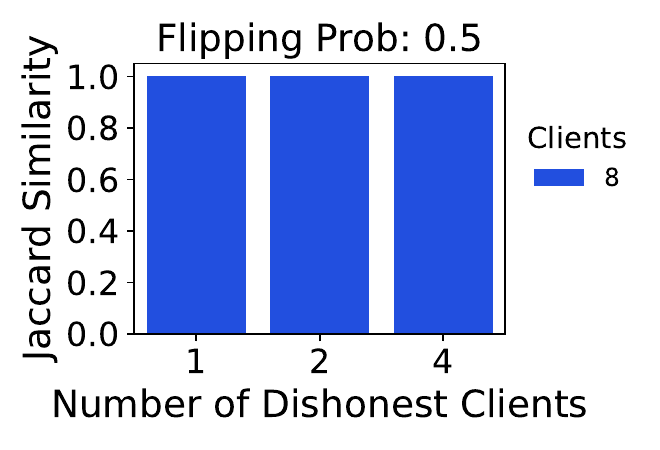}
	\includegraphics[scale=0.4]{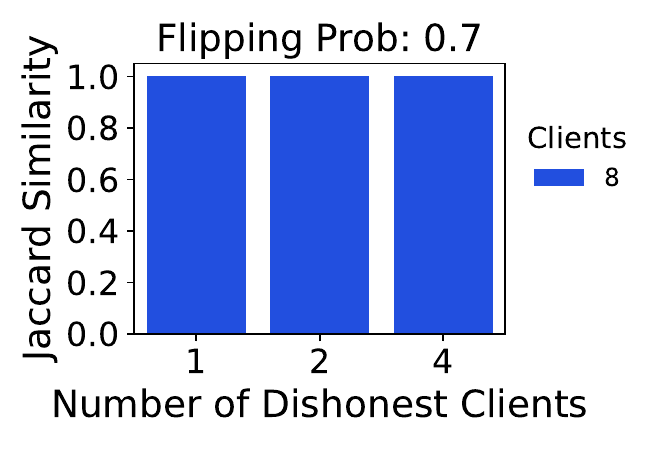}
	\includegraphics[scale=0.4]{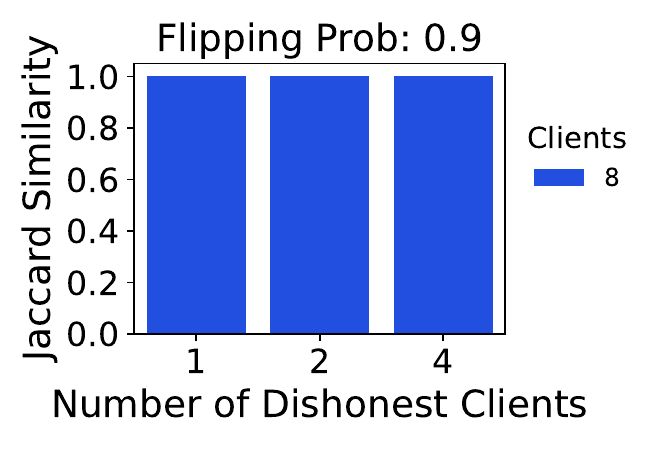}
	\caption{Effectiveness of separating honest clients while varying dishonest clients. A higher Jaccard similarity denotes a higher separation. We consider flipping window size $ 5 $. In each epoch, $ 100\% $ clients participate in training.}
\end{figure*}

\begin{figure*}
	\centering
	\includegraphics[scale=0.4]{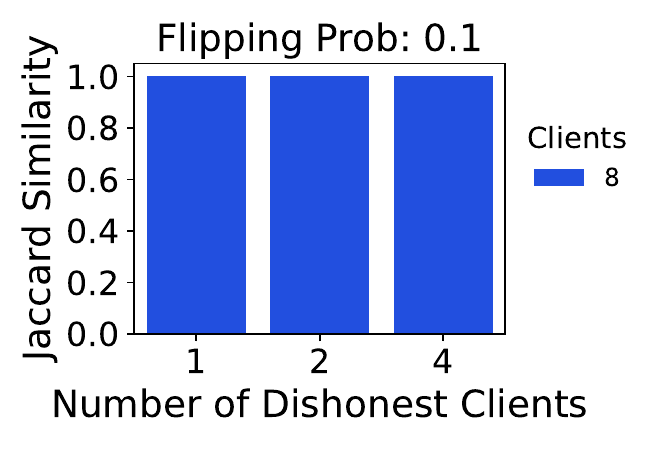}
	\includegraphics[scale=0.4]{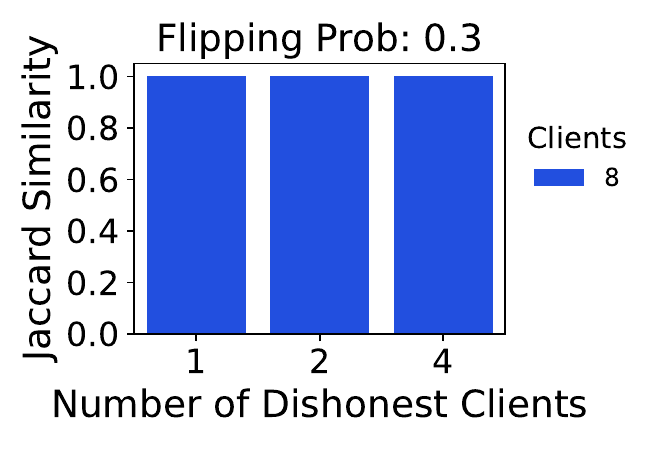}	
	\includegraphics[scale=0.4]{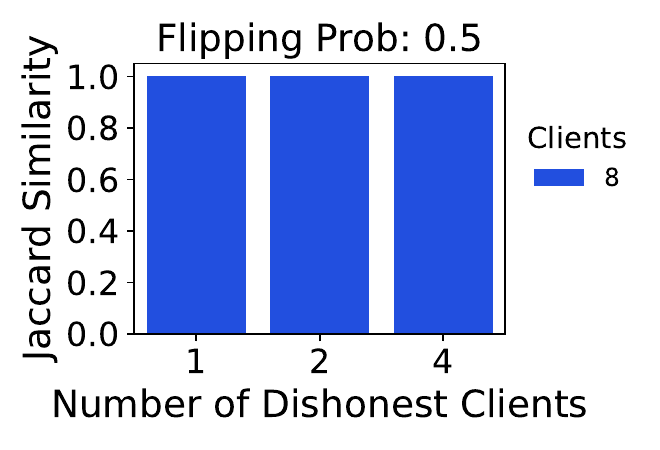}
	\includegraphics[scale=0.4]{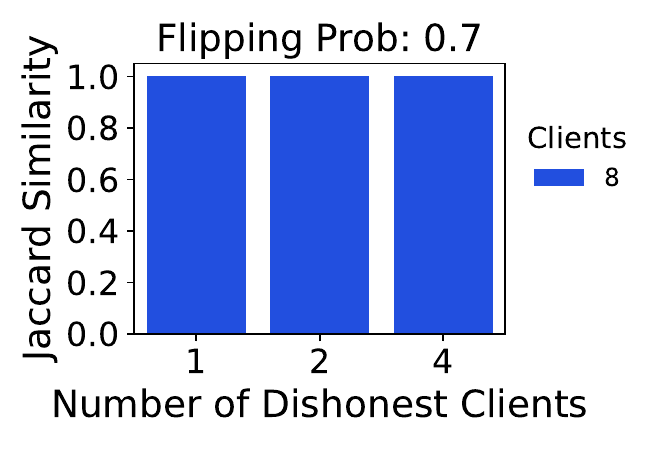}
	\includegraphics[scale=0.4]{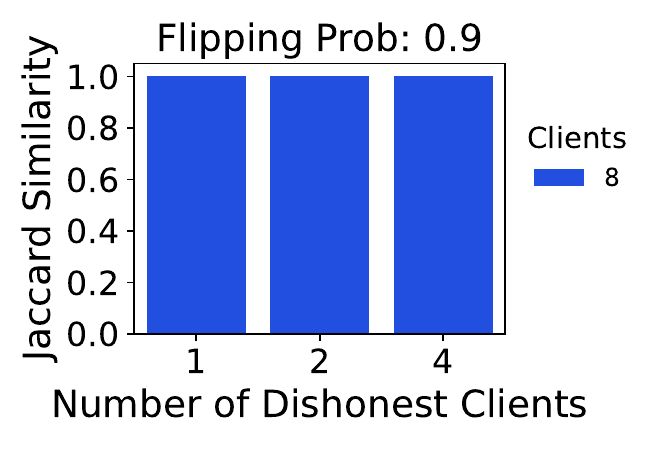}
	\caption{Effectiveness of separating honest clients while varying dishonest clients. A higher Jaccard similarity denotes a higher separation. We consider flipping window size $ 10 $. In each epoch, $ 100\% $ clients participate in training.}
\end{figure*}

\clearpage

\begin{figure*}[t!]
	\centering
	\includegraphics[scale=0.4]{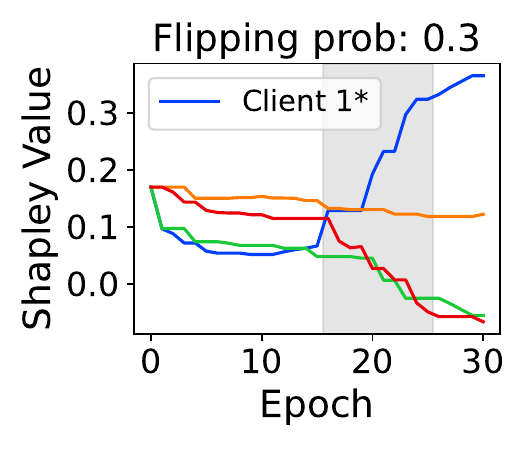}
	\includegraphics[scale=0.4]{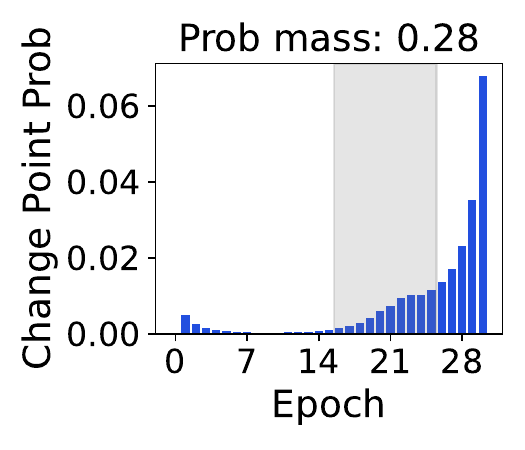}
	\includegraphics[scale=0.4]{figures/client_intention_analysis/client_intent_analysis_shapley_value_4_1_0.5_0.5_0_10.pdf}
	\includegraphics[scale=0.4]{figures/client_intention_analysis/client_intent_analysis_change_point_loss_4_1_0.5_0.5_0_10.pdf}
	\includegraphics[scale=0.4]{figures/client_intention_analysis/client_intent_analysis_shapley_value_4_1_0.5_0.7_0_10.pdf}
	\includegraphics[scale=0.4]{figures/client_intention_analysis/client_intent_analysis_change_point_loss_4_1_0.5_0.7_0_10.pdf}

	\includegraphics[scale=0.4]{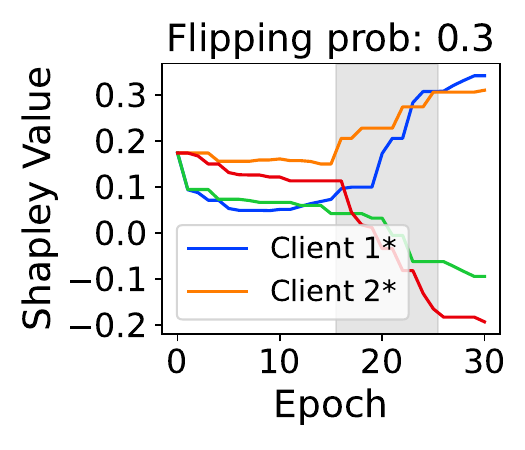}
	\includegraphics[scale=0.4]{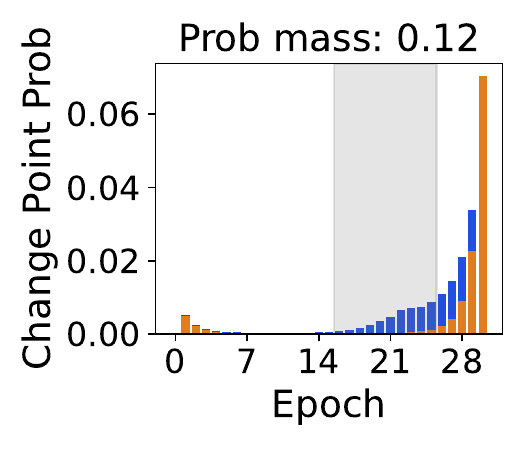}
	\includegraphics[scale=0.4]{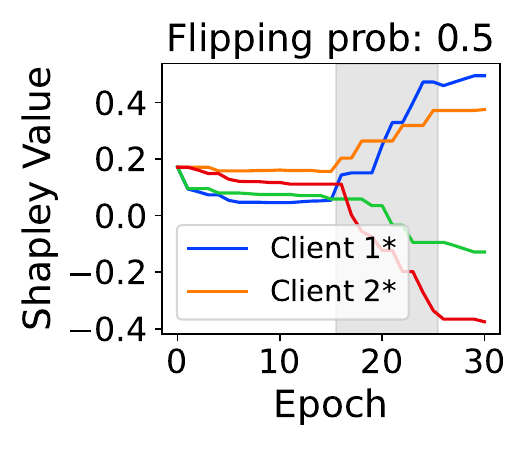}
	\includegraphics[scale=0.4]{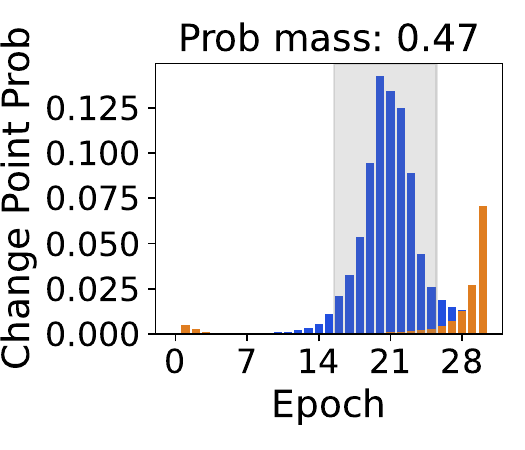}
	\includegraphics[scale=0.4]{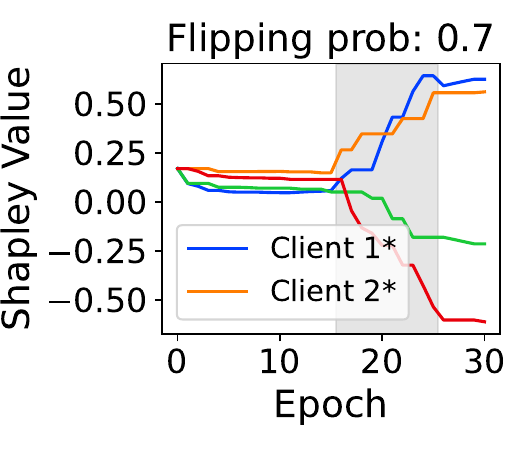}
	\includegraphics[scale=0.4]{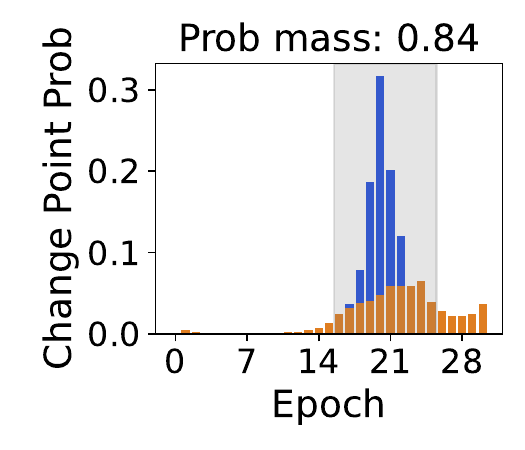}
	
	\caption{Cumulative Shapley values and corresponding change point detection probabilities of four clients with one (top two rows) and two (bottom two rows) dishonest clients. Gray area denotes the poisonous window.}
\end{figure*}

\begin{figure*}[t!]
	\centering
	\includegraphics[scale=0.4]{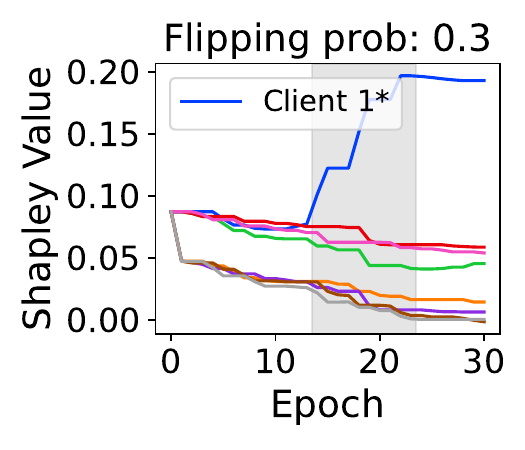}
	\includegraphics[scale=0.4]{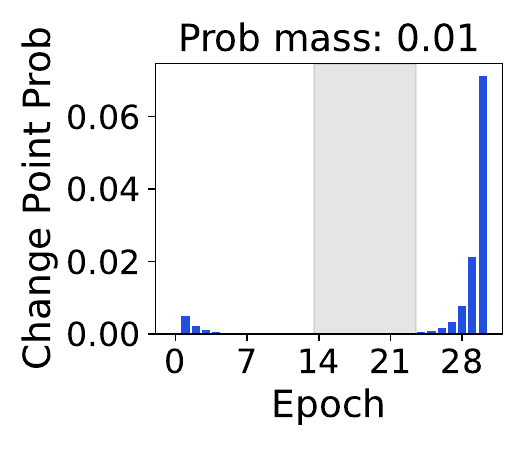}
	\includegraphics[scale=0.4]{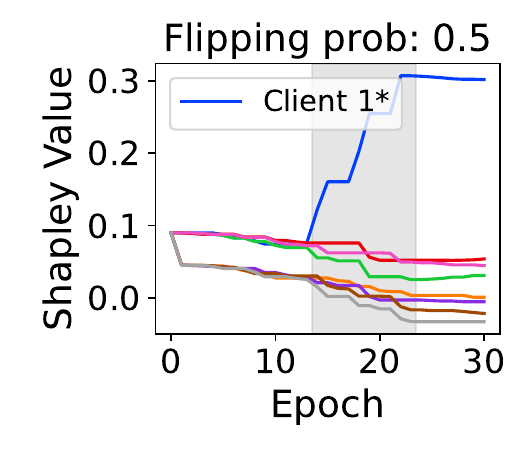}
	\includegraphics[scale=0.4]{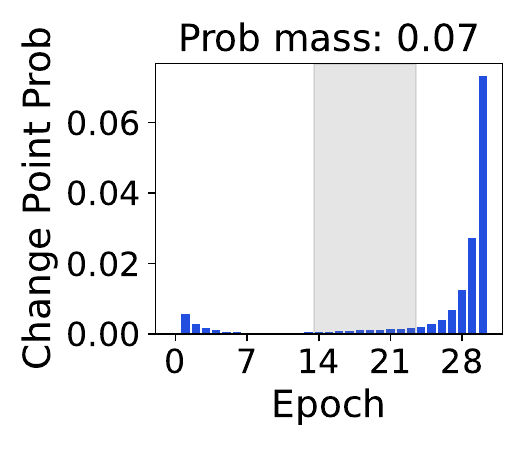}
	\includegraphics[scale=0.4]{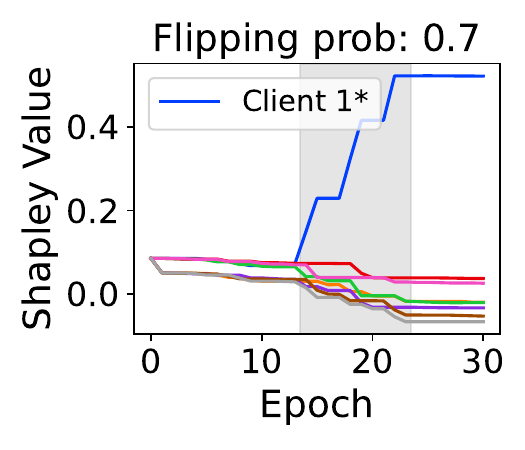}
	\includegraphics[scale=0.4]{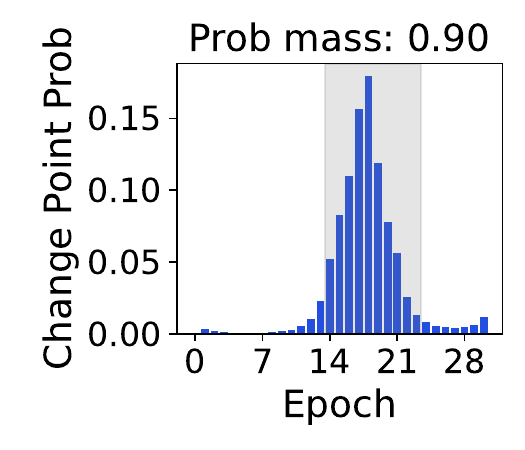}

	\includegraphics[scale=0.4]{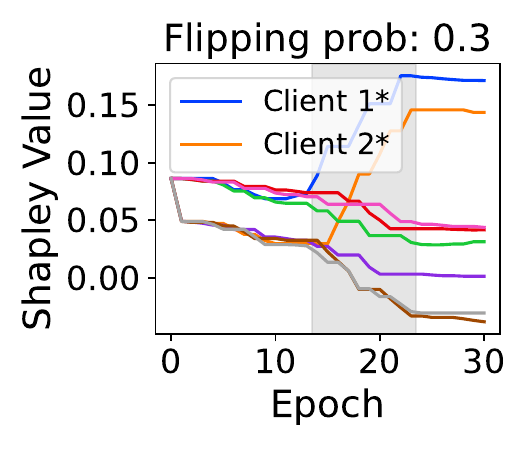}
	\includegraphics[scale=0.4]{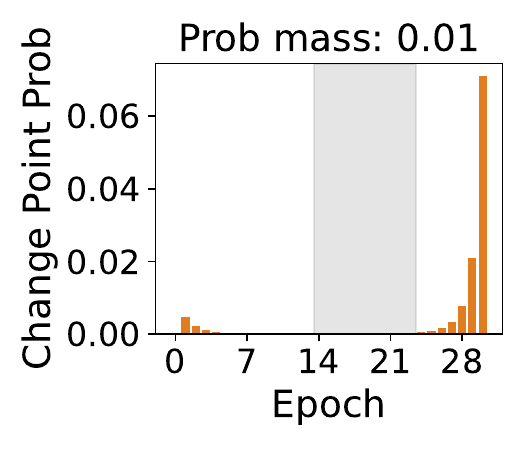}
	\includegraphics[scale=0.4]{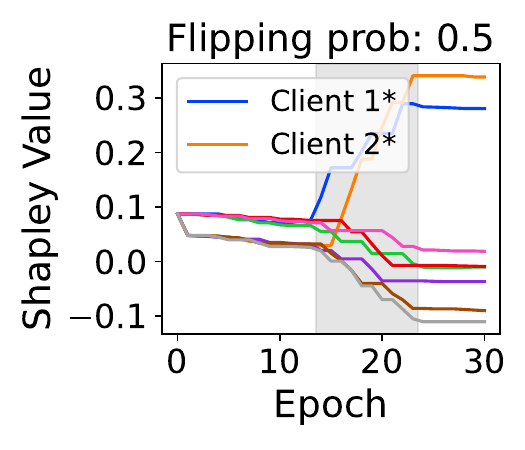}
	\includegraphics[scale=0.4]{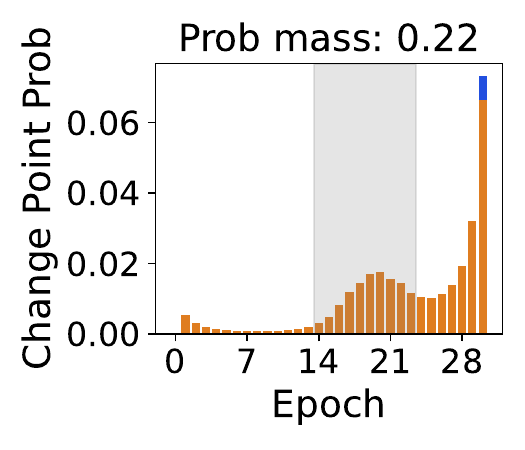}
	\includegraphics[scale=0.4]{figures/client_intention_analysis/client_intent_analysis_shapley_value_8_2_0.5_0.7_0_10.pdf}
	\includegraphics[scale=0.4]{figures/client_intention_analysis/client_intent_analysis_change_point_loss_8_2_0.5_0.7_0_10.pdf}
	
	\caption{Cumulative Shapley values and corresponding change point detection probabilities of eight clients with one (top two rows) and two (bottom two rows) dishonest clients. Gray area denotes the poisonous window.}
\end{figure*}

\begin{figure*}
	\centering
	\includegraphics[scale=0.4]{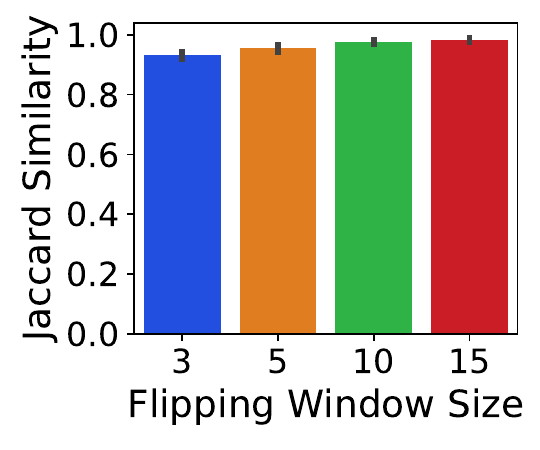}
	\caption{Effectiveness of separating honest clients for varying flipping window. A higher Jaccard similarity denotes a higher separation}
\end{figure*}

\end{document}